\pdfoutput=1
\documentclass[lettersize,journal]{IEEEtran}
\usepackage[utf8]{inputenc}
\usepackage{amsmath,amsfonts,amssymb}
\usepackage{amsthm}
\usepackage{algorithmic}
\usepackage{algorithm}
\usepackage{array}
\usepackage[caption=false,font=normalsize,labelfont=sf,textfont=sf]{subfig}
\usepackage{textcomp}
\usepackage{stfloats}
\usepackage[hyphens]{url}
\usepackage{verbatim}
\usepackage{graphicx}
\usepackage{cite}
\usepackage{booktabs}
\usepackage{enumitem}
\usepackage{diagbox}
\usepackage{makecell}
\usepackage{xcolor,colortbl}
\usepackage{multirow}
\usepackage{bm}
\usepackage[hidelinks]{hyperref}

\newcommand{\tablefontsize}[0]{%
\fontsize{7pt}{9pt}\selectfont
}
\newcommand{\tablefontsizeXS}[0]{%
\fontsize{7.0pt}{9pt}\selectfont
}
\definecolor{Gray}{gray}{0.93}
\definecolor{revisioncolor}{HTML}{000000}
\newcommand{\revision}[1]{{\color{revisioncolor}#1}}
\newenvironment{theoryrevision}{\color{black}}{}
\newcommand{\greenrevision}[1]{\textcolor{black}{#1}}
\definecolor{theorypurple}{HTML}{000000}
\newcommand{\purplerevision}[1]{{\color{theorypurple}#1}}
\newenvironment{purplerevisionblock}{\color{theorypurple}}{}
\definecolor{lowrankgreen}{HTML}{000000}

\newenvironment{lowrankrevisionblock}{\color{lowrankgreen}}{}
\definecolor{secondlowrankpurple}{HTML}{000000}

\newenvironment{secondlowrankrevisionblock}{\color{secondlowrankpurple}}{}

\newcommand{\R}{\mathbb{R}}
\DeclareMathOperator{\vecop}{vec}
\DeclareMathOperator*{\argmax}{arg\,max}

\newtheorem{theorem}{Theorem}
\newtheorem{corollary}{Corollary}

\begin{document}

\title{LoRAP: Low-Rank Aggregation Prompting for Quantized Graph Neural Networks Training}
\author{Chenyu Liu, Haige Li, and Luca Rossi
\thanks{Chenyu Liu and Luca Rossi are with the Department of Electrical and Electronic Engineering, The Hong Kong Polytechnic University, Hong Kong SAR, China (e-mail: chen-yu.liu@connect.polyu.hk; luca.rossi@polyu.edu.hk).}
\thanks{Haige Li is with the Department of Data Science and Artificial Intelligence, The Hong Kong Polytechnic University, Hong Kong SAR, China.}
\thanks{Corresponding author: Luca Rossi.}}

\markboth{IEEE Transactions on Knowledge and Data Engineering}%
{Liu \MakeLowercase{\textit{et al.}}: LoRAP: Low-Rank Aggregation Prompting for Quantized Graph Neural Networks Training}

\maketitle

\begin{abstract}
Graph Neural Networks (GNNs) are neural networks that aim to process graph data, capturing the relationships and interactions between nodes using the message-passing mechanism. GNN quantization has emerged as a promising approach for reducing model size and accelerating inference in resource-constrained environments. Compared to quantization in LLMs, quantizing graph features is more emphasized in GNNs. Inspired by the above, we propose to leverage prompt learning, which manipulates the input data, to improve the performance of quantization-aware training (QAT) for GNNs. \purplerevision{To directly compensate for aggregation-level quantization errors, we introduce Low-Rank Aggregation Prompting (LoRAP), which injects lightweight, input-dependent prompts into the aggregated representations. Our analysis characterizes the optimization geometry and locality of this insertion position and gives conditions under which the correction benefit is preserved after requantization.} Extensive evaluations on 4 leading QAT frameworks over \purplerevision{10} graph datasets demonstrate that LoRAP consistently enhances the performance of low-bit quantized GNNs while introducing a minimal computational overhead.

\end{abstract}


\section{Introduction}
\IEEEPARstart{G}{raph} Neural Networks (GNNs) are models able to process graph data, effectively capturing the node relationships using the message-passing mechanism \cite{zhang2022deep}. This ability allows GNNs to model complex dependencies and interactions present in graphs and enabled their application in several domains, such as biology~\cite{hetzel2021graph}, chemistry~\cite{reiser2022graph,minello2025graph}, recommender systems ~\cite{wu2022graph}, and network analysis~\cite{tang2010graph,rahmani2023graph}. Despite their wide use, the adoption of GNNs in resource-constrained environments, such as mobile phones, autonomous vehicles, and edge devices, is still challenging \cite{wu2022graph}. Even though they have fewer parameters than standard deep learning models, GNNs typically involve more computations, often exceeding the computational cost of standard models \cite{tailor2020degree,ma2025acceleration}. This, in turn, is largely driven by the size of the underlying graph.

When dealing with large neural networks, particularly large language models (LLMs), model quantization has emerged as a promising approach to handle their deployment in resource-constrained environments~\cite{dettmers2022gpt3}.
This in turn involves representing weights and activations with lower precision, reducing the model size and accelerating inference.

\begin{figure}[!t]
\centering
\includegraphics[width=0.8\linewidth]{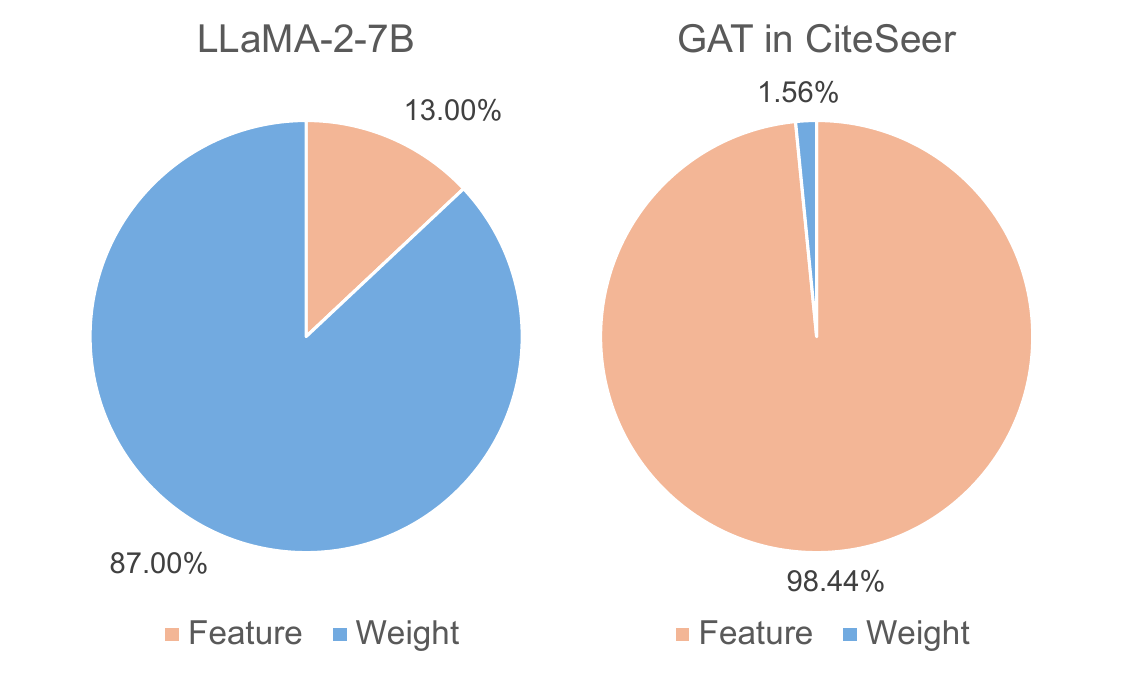}
\caption{Feature/Weight memory size ratio during inference in LLMs vs GNNs.}
\label{memory}
\end{figure}

\begin{figure*}[!t]
\centering
\includegraphics[width=0.7\linewidth]{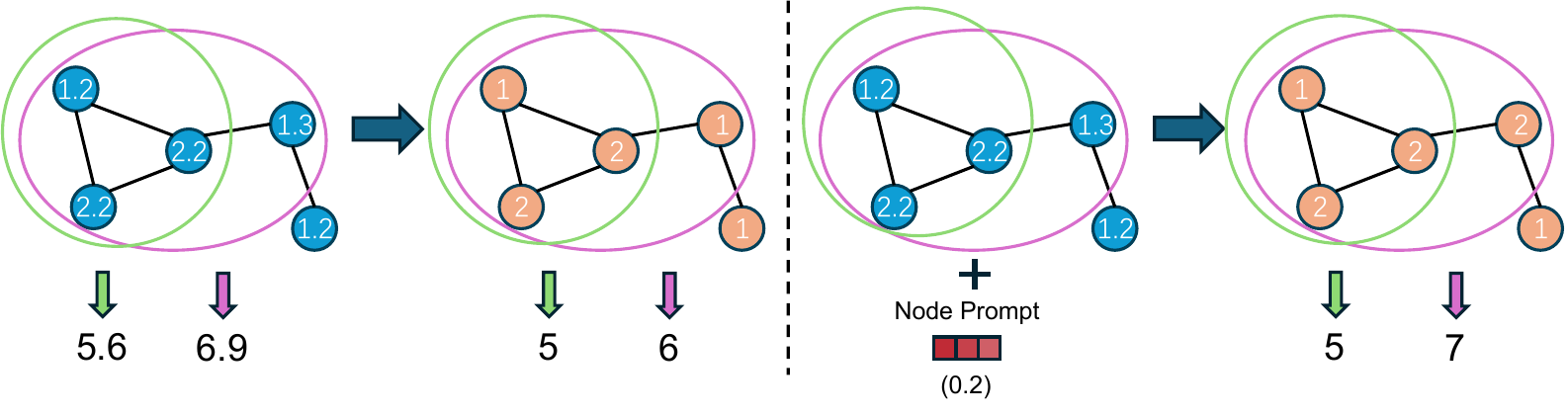}
\caption{(Left) The aggregation phase in a full-precision GNN and its quantized version.
(Right) The aggregation phase in a full-precision prompt-tuned GNN (where a node prompt with a value of 0.2 is added) and its quantized version.
The neighbourhoods where the aggregations take place are denoted by green \& magenta circles, respectively, with the aggregated values indicated by green \& magenta arrows, respectively.}
\label{fig:motivation1}
\end{figure*}

While traditional quantization techniques focus primarily on the weights of the model, in the case of GNNs the quantization of node features and graph structure becomes significantly more important.
As the graph scale increases, the necessary storage size escalates dramatically, severely impacting the limited memory budget of resource-constrained devices.
In contrast, memory consumption in LLMs is overwhelmingly dominated by parameter weights.
As shown in Fig. \ref{memory},
Feature representations can comprise up to 98.44\% of the total memory footprint of a GNN~\cite{feng2020sgquant}, whereas in LLaMA-2-7B roughly 87\% of the overall memory usage is attributable to model weights~\cite{li2023qft}.
Note also that GNN inference involves repeated message passing over node embeddings, thus any reduction in the bit-width of node data directly shrinks memory transfers and arithmetic operations on each edge, yielding substantial speedups and bandwidth savings in real-world deployments.

The reduced memory footprint and runtime, however, usually result in decreased performance on downstream tasks.
This drop is particularly pronounced when quantization is applied to GNNs, where the relationships between nodes and their neighbors are crucial for effective learning.
For this reason, several approaches have been introduced to incorporate quantization-aware training (QAT) for GNNs.
These methods try to mitigate quantization errors by leveraging node degree information \cite{tailor2020degree} or using mixed-precision strategies \cite{zhu2023rm}.
Despite this, quantizing GNNs to low-bit representations still incurs a large accuracy drop.

\purplerevision{Since feature quantization errors propagate through graph message passing,} in this paper we propose to use prompt learning, a technique where additional learnable parameters are injected into data features, to mitigate the performance degradation of quantized GNNs.
Prompt tuning was introduced as a technique to recover the performance of compressed LLMs \cite{xu2024soft}, with most methods in this context being adapter-based \cite{dettmers2024qlora}, since quantization in LLMs emphasizes weights. However, extending this approach to GNN models is not straightforward and incurs considerable challenges \cite{fang2024universal}.

To this end, in this paper we first propose to recover the performance of quantized GNNs by incorporating prompts in the node features space.
We implement GPF-plus \cite{fang2024universal} for QAT, which integrates different prompted features for each node in the graph.
\purplerevision{However, pre-aggregation prompts are transformed by neighborhood aggregation, coupling their effects across receivers, as illustrated in Fig.~\ref{fig:motivation1} (Right) for a shared node prompt.}
\purplerevision{We therefore insert prompts directly into the aggregation outputs, where each prompt vector acts on its corresponding receiver and compensates for the aggregation residual.}
Our prompts are also designed to be input-dependent and computationally efficient, making use of low-rank basis matrices.

Our contributions can be summarized as follows:
\begin{itemize}[leftmargin=*]
\item We are the first to leverage prompt learning to improve the performance of quantized GNN models. Specifically, we manipulate the node features, the major source of reduced performance in quantized GNNs.
\item We propose LoRAP, a \textbf{Low-Rank Aggregation Prompting} strategy where we insert input-dependent computationally efficient prompts over node feature aggregations, improving the quantized aggregation results; Moreover, to reduce the latency, we propose to fuse the GPU operations in LoRAP, resulting in the fused LoRAP kernel.
\item Experiments on 4 QAT frameworks for GNNs over a wide variety of datasets demonstrate that combining GPF-plus and LoRAP consistently increases the model performance.
\end{itemize}

\section{Background and Related Work}
\subsection{Quantization Process.}
\label{appendix:quantization}
Given a real-valued tensor $x$ and bit-width $b$, we define the \emph{quantization operator} $Q(\cdot)$ and \emph{dequantization operator} $DQ(\cdot)$ as
\begin{equation}
Q(x) = \mathrm{clip}\Bigl(\mathrm{round}\bigl(\tfrac{x}{\purplerevision{\Delta}}\bigr) + \purplerevision{z_0},\;q_{\min},\;q_{\max}\Bigr)
\end{equation}
\begin{equation}
DQ(x_q) = \purplerevision{\Delta}\,(x_q - \purplerevision{z_0}) \,,
\end{equation}
\purplerevision{where, for a non-degenerate range $x_{\max}>x_{\min}$, the positive scale $\Delta$ and zero-point $z_0$ are computed as}
\[
\purplerevision{\Delta} = \frac{x_{\max} - x_{\min}}{q_{\max}-q_{\min}},
\qquad
\purplerevision{z_0 = q_{\min}-\mathrm{round}\!\Bigl(\frac{x_{\min}}{\Delta}\Bigr),}
\]
and the integer quantization bounds are
\[
q_{\min} = 0,\quad q_{\max} = 2^b - 1
\quad(\text{unsigned}),
\]
or
\[
q_{\min} = -2^{\,b-1},\quad q_{\max} = 2^{\,b-1}-1
\quad(\text{signed}).
\]
Here, $x_{\min}$ and $x_{\max}$ denote the minimum and maximum values of $x$. The function $\mathrm{round}(\cdot)$ maps its argument to the nearest integer, and $\mathrm{clip}(\cdot,q_{\min},q_{\max})$ truncates values outside $[q_{\min},q_{\max}]$ to the nearest bound. During inference, each real value $x$ is quantized to $x_q = Q(x)$ and then dequantized back to $\hat x = DQ(x_q)$ for low-bitwidth storage and arithmetic.

\subsection{Quantization Techniques.}
\label{sec:quantization-techniques}
Quantization is a key strategy for reducing the memory footprint and computational cost of deep neural networks by representing weights and activations with lower-precision numbers. Early work showed that 16-bit fixed-point quantization can be applied to training and inference with minimal accuracy loss and was later extended to 8-bit and sub-8-bit precisions \cite{jacob2018quantization}. Broadly speaking, these methods fall into two categories:

\begin{itemize}
\item\textbf{Post-Training Quantization (PTQ).} PTQ methods perform a lightweight calibration step on a pretrained full-precision model to determine optimal quantization parameters (namely scale, zero-point, and rounding directions) using only a small calibration dataset. A simple min-max calibration maps the observed activation range to integer bounds, while histogram-based techniques clip outliers by optimizing for minimal information loss according to percentile criteria \cite{nagel2020adaptive}.

\item\textbf{Quantization-Aware Training (QAT).} QAT injects fake-quantization operations into both forward and backward passes so that the network learns to adapt its weights to low-precision representations. Early QAT works on binary and ternary networks demonstrated the feasibility of end-to-end gradient updates under discrete constraints \cite{hubara2017quantized}. Later methods introduced learnable scale and zero-point parameters \cite{jacob2018quantization}, while Learned Step Size Quantization (LSQ) makes the quantization grid itself trainable for optimal precision-range trade-offs \cite{esser2020learned}. Unlike PTQ, QAT updates both weights and quantization parameters during training, yielding higher accuracy, especially at very low bit-widths, at the cost of significantly longer training times.
\end{itemize}

\subsection{Prompt Tuning.}
Prompt tuning has emerged as a highly parameter-efficient strategy for adapting large pretrained models to downstream tasks by learning a small set of continuous prompt vectors, while keeping the original model weights frozen. Early work on in-context learning with discrete prompts demonstrated the power of prompting for few-shot tasks \cite{brown2020language}. Building on this idea, Lester et al.\ \cite{lester2021power} introduced \emph{Prompt Tuning}, in which soft prompt embeddings are prepended to the input and optimized via gradient descent, achieving performance competitive with full fine-tuning using orders of magnitude fewer trainable parameters. Li and Liang \cite{li2021prefix} further generalized this to \emph{Prefix Tuning}, where trainable vectors are injected at every layer of a transformer. Liu et al.\ \cite{liu2021ptuning,liu2022ptuning} proposed \emph{P-Tuning} and \emph{P-Tuning v2}, which scale soft prompting to very large language models by optimizing continuous prompts at multiple layers. In the vision domain, Visual Prompt Tuning (VPT) \cite{jia2022visual} applies a similar soft-prompt mechanism to vision transformers. Despite these advances in natural language processing and computer vision, prompt tuning has been largely unexplored for graph-based models. Our work is the first to explore the use of prompt tuning for GNNs QAT.

{

\subsection{Graph Prompting.}
Inspired by the success of prompt learning in Natural Language Processing (NLP), graph prompting has emerged as a paradigm to bridge the gap between pre-training and downstream tasks by modifying the input graph rather than fine-tuning the entire model parameters.
Early works focused on input-space adaptation: GPF and GPF-Plus~\cite{fang2024universal} introduce learnable vectors to node features, while GraphPrompt+~\cite{yu2024generalized} unifies pre-training and downstream graph tasks through a shared template and inserts learnable prompt vectors across multiple GNN layers. Graph-augmented prompt tuning has also been explored for low-resource text classification~\cite{wen2024prompt}.
More recently, unified frameworks like UniPrompt~\cite{huangone} and EdgePrompt~\cite{fuedge} have attempted to optimize feature and structure prompts simultaneously.
Theoretical understanding of this paradigm has also deepened; Wang et al.~\cite{wang2025does} analyzed graph prompting from a data operation perspective, providing rigorous error bounds and proving that prompts essentially simulate graph transformation operators to align upstream and downstream tasks.
Furthermore, to address distribution shifts in complex scenarios such as heterophilic graphs, DAGPrompT~\cite{chen2025dagprompt} proposes a distribution-aware tuning framework that customizes prompts for hop-specific node requirements, significantly pushing the limits of prompting effectiveness.

\subsection{Parameter-Efficient Fine-Tuning and LoRA on Graphs.}
Parallel to prompting, Low-Rank Adaptation (LoRA) has gained traction for efficient model adaptation. In the graph domain, GraphLoRA~\cite{graphlora2024} applies LoRA to GNN weights combined with contrastive learning to enable efficient cross-graph transfer learning without catastrophic forgetting. Similarly, the GLoRA module within DAGPrompT~\cite{chen2025dagprompt} utilizes low-rank adaptation to optimize the GNN encoder's projection matrices for better distribution alignment.
\textit{Difference from our work:} It is crucial to distinguish our proposed LoRAP from these approaches. While GraphLoRA and DAGPrompT employ LoRA to adapt the \textit{model backbone weights} for transfer learning tasks, our method applies a low-rank constraint specifically to the \textit{aggregation prompt} injected into the message-passing stream. To the best of our knowledge, our work is the first to leverage low-rank prompting specifically as a compensation mechanism for \textit{Quantization-Aware Training (QAT)}, \purplerevision{using low-rank basis matrices to parameterize input-dependent corrections with fewer trainable parameters.}
}
\subsection{Quantization in GNNs.}
Graph Neural Network (GNN) quantization has recently attracted attention as a means to reduce model size and accelerate inference on graph data \cite{zhang2026survey,ma2025acceleration}, complementing scalable graph sampling approaches~\cite{ding2025scalable}. Unlike CNNs, GNNs pose unique challenges for quantization. Most notably, the irregular graph topology leads to highly variable node activation magnitudes (due to differing neighbor counts), making naive, uniform low-bit quantization prone to severe accuracy loss \cite{tailor2020degree}. Early attempts to apply standard 8-bit quantization to GNNs often resulted in drastic performance degradation at lower bit-widths \cite{wang2020binarized}. To address these issues, researchers have developed degree-aware, adaptive-precision, and other specialized quantization frameworks that explicitly leverage graph structure. Below, we summarize key approaches and their innovations in the context of GNN quantization.

Recent advances in GNN quantization emphasize leveraging graph structure to guide precision allocation. Degree-Quant (DQ) \cite{tailor2020degree} introduces a stochastic masking scheme during QAT that ``protects'' high-degree nodes in full precision and employs percentile-based clipping to determine optimal scale ranges, enabling both 4-bit and 8-bit models to nearly match full-precision accuracy. {A\textsuperscript{2}Q} \cite{zhu2023rm} further augments this idea by learning a mixed-precision assignment for each node according to its aggregation complexity: low-degree nodes use very few bits, while high-degree ``hub'' nodes receive higher bit-width. Continuous proxies and gradient-friendly approximations allow per-node bit optimization. In conclusion, these graph-aware and adaptive quantization strategies deliver efficient yet high-accuracy GNNs suitable for both large-scale and edge deployments.

\section{Fundamentals of Graph Neural Networks}
Unlike traditional convolutional neural networks that operate on grid-like data, GNNs leverage the graph topology to propagate information across nodes, enabling the model to learn representations that capture the relational inductive biases inherent in graphs.
Consider a graph \( \mathcal{G} = (\mathcal{V}, \mathcal{E}) \) consisting of a set of nodes \( \mathcal{V} = \{v_1, v_2, \dots, v_N\} \) and edges \( \mathcal{E} \), where each node \( v_i \) is associated with a feature vector \( \mathbf{x}_i \in \mathbb{R}^d \).
Let \( \mathbf{X} = [\mathbf{x}_1; \mathbf{x}_2; \dots; \mathbf{x}_N] \in \mathbb{R}^{N \times d} \) the matrix of these features and \( \mathbf{A} \in \mathbb{R}^{N \times N} \) be the adjacency matrix encoding the connectivity between nodes, where \( \mathbf{A}_{ij} \) denotes the presence (and possibly the weight) of an edge between nodes \( v_i \) and \( v_j \).
The message passing mechanism at the core of GNNs consists of three key phases: message passing, aggregation, and node update.
In the message passing phase, each node \(v_i\) sends a message to its neighbors based on its current state and edge information.
The message from node \(v_i\) to its neighbor \(v_j\) is computed as a function \( \phi \) that depends on the node features and the edge features, \emph{i.e.},
\[
\mathbf{m}_{ij} = \phi \left( \mathbf{h}_i^{(l)}, \mathbf{h}_j^{(l)}, \mathbf{e}_{ij} \right) \,,
\]
where \( \mathbf{h}_i^{(l)} \) and \( \mathbf{h}_j^{(l)} \) are the features of \(v_i\) and \(v_j\) at layer \(l\) and \( \mathbf{e}_{ij} \) represents the edge feature between \(v_i\) and \(v_j\).
In the aggregation phase, the messages received from neighbors are combined into a single message for each node.
The aggregation function \( \bigwedge \) is typically permutation-invariant, ensuring that the order of neighbors does not affect the result.
The aggregated message for node \(v_i\) can be expressed as
\[
\mathbf{s}_i = \bigwedge_{j \in \mathcal{N}(i)} \mathbf{m}_{ij} = \bigwedge_{j \in \mathcal{N}(i)} \phi \left( \mathbf{h}_i^{(l)}, \mathbf{h}_j^{(l)}, \mathbf{e}_{ij} \right) \,,
\]
where \( \mathcal{N}(i) \) is the set of neighbors of node \(v_i\), and \( \bigwedge \) can be a sum, mean, or max function, depending on the architecture.
Each node finally updates its representation based on its current features and the aggregated messages, typically using a learnable function \( \gamma \), such as a multi-layer perceptron.
The update rule for node \(v_i\) at layer \(l+1\) is
\[
\mathbf{h}_i^{(l+1)} = \gamma \left( \mathbf{h}_i^{(l)}, \mathbf{s}_i \right) \,.
\]

\begin{figure*}[!t]
\centering
\includegraphics[width=0.6\linewidth]{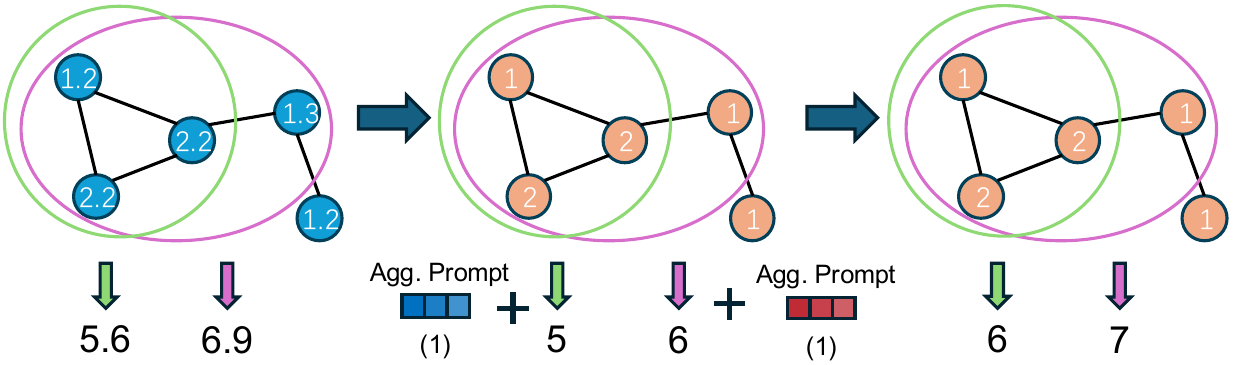}
\caption{\purplerevision{The aggregation prompting strategy. Input-dependent prompts are added to the quantized aggregation outputs to compensate for their residuals relative to the full-precision representations.}}
\label{fig:motivation2}
\end{figure*}
\begin{purplerevisionblock}
\section{Prompt Tuning For Quantized GNNs}
\label{sec:prompt-theory}

\subsection{Quantization and reconstruction error}
Following Section~\ref{appendix:quantization}, $Q$ maps real-valued tensors to integer codes and $DQ$ maps those codes back to real units. For node features $X\in\mathbb{R}^{N\times d}$, write
\[
X_q=DQ(Q(X)),\qquad \epsilon_X=X_q-X.
\]
Each pair $Q,DQ$ uses the same scale and zero-point; different tensors may use different quantization parameters. The feature reconstruction loss is
\begin{equation}
\mathcal L_{\mathrm{feat}}=\tfrac12\|X-X_q\|_F^2.
\label{eq:feature_loss}
\end{equation}
Feature reconstruction is a common quantization objective~\cite{zhu2023rm}, but aggregation transforms these feature errors through the graph. Consequently, a small feature error need not produce an equally small aggregation error.

At the same layer input and model checkpoint, let $S_{\mathrm{fp}}\in\mathbb{R}^{N\times d}$ be the full-precision aggregation output and $\widehat S\in\mathbb{R}^{N\times d}$ the dequantized aggregation output before prompt insertion. Define
\begin{equation}
R=S_{\mathrm{fp}}-\widehat S,\qquad
\mathcal L_{\mathrm{agg}}=\tfrac12\|R\|_F^2.
\label{eq:aggregation_loss}
\end{equation}
For a continuous prompt output $P\in\mathbb{R}^{N\times d}$, the corrected representation and its remaining error are
\begin{equation}
Z_P=\widehat S+P,\qquad
\varepsilon(P)=\|S_{\mathrm{fp}}-Z_P\|_F=\|R-P\|_F.
\label{eq:prompt-residual}
\end{equation}
These reconstruction losses describe the correction mechanism. LoRAP is trained with the original QAT objective, as described in Section~\ref{sec:joint-qat}.

\subsection{Post-aggregation correction and optimization geometry}
Prompting introduces learnable adjustments that can bring quantized representations closer to their full-precision references. A feature-level example is GPF~\cite{fang2024universal}, which adds a shared vector $p\in\mathbb{R}^{d}$ to every node. In Fig.~\ref{fig:motivation1}, the same scalar adjustment improves one neighborhood's quantized aggregation but does not simultaneously correct the other, illustrating the restriction of a shared prompt across distinct neighborhoods; GPF-plus instead generates node-dependent prompts.

Aggregation prompting places the correction directly at the aggregation output. To isolate the effect of this placement, fix a linear aggregation operator $A\in\mathbb{R}^{N\times N}$ and the input features, and consider
\[
\begin{aligned}
S_{\mathrm{fp}}&=AX,\qquad \widehat S=AX_q,\\
S_{\mathrm{post}}(P)&=AX_q+P,\\
S_{\mathrm{pre}}(P)&=A(X_q+P).
\end{aligned}
\]
Here $R=-A\epsilon_X$, and the pre-aggregation model $S_{\mathrm{pre}}(P)$ inserts $P$ after input quantization (distinct from $A\,DQ(Q(X+P))$).

\begin{theorem}[Conditional optimization geometry]
\label{thm:optimization-geometry}
Under the fixed-input linear model above, define
\begin{equation}
\begin{aligned}
\mathcal L_{\mathrm{post}}(P)&=\tfrac12\|R-P\|_F^2,\\
\mathcal L_{\mathrm{pre}}(P)&=\tfrac12\|R-AP\|_F^2.
\end{aligned}
\label{eq:prompt-objectives}
\end{equation}
Their gradients and Hessians in column-vectorized prompt-output coordinates are
\begin{equation}
\begin{aligned}
\nabla_P\mathcal L_{\mathrm{post}}&=P-R,\\
\nabla_P\mathcal L_{\mathrm{pre}}&=A^\top(AP-R),\\
\nabla^2\mathcal L_{\mathrm{post}}&=I_{Nd},\\
\nabla^2\mathcal L_{\mathrm{pre}}&=I_d\otimes A^\top A.
\end{aligned}
\label{eq:prompt-hessians}
\end{equation}
The post-aggregation condition number is $1$. For $A\ne0$, the pre-aggregation condition number on its nonzero-curvature subspace is
\begin{equation}
\kappa_{\mathrm{pre},+}
=\frac{\sigma_{\max}^2(A)}{\sigma_{\min,+}^2(A)},
\label{eq:prompt-condition}
\end{equation}
where $\sigma_{\min,+}(A)$ is the smallest positive singular value.
\end{theorem}

\begin{proof}
Differentiating the two quadratic objectives gives the gradients and the Hessian operators $H\mapsto H$ and $H\mapsto A^\top AH$. Stacking matrix columns yields the stated Hessians. The positive eigenvalues of $A^\top A$ are the squared positive singular values of $A$, which gives the condition number. The full derivation and residual dynamics appear in Appendix~A.1.
\end{proof}

Post-aggregation placement removes the curvature scaling induced by $A$. A singular $A$ introduces flat directions $AH=0$ in the pre-aggregation objective, while unequal positive singular values introduce curvature anisotropy.

\subsection{Locality of correction outputs}
The same model also characterizes how a node's prompt affects the receiving nodes and their reconstruction gradients.

\begin{theorem}[Insertion-level locality and targeted correction]
\label{thm:insertion-locality}
For fixed $A$ and $X_q$, let $P_j\in\mathbb{R}^{1\times d}$ denote the prompt vector associated with node $j$, namely the $j$th row of the free matrix $P$. Then
\begin{equation}
\begin{aligned}
\frac{\partial(S_{\mathrm{post}}(P))_i}{\partial P_j}
&=\delta_{ij}I_d,\\
\frac{\partial(S_{\mathrm{pre}}(P))_i}{\partial P_j}
&=A_{ij}I_d,
\end{aligned}
\label{eq:insertion-locality}
\end{equation}
where $\delta_{ij}$ is the Kronecker delta. Under \eqref{eq:prompt-objectives}, the node-level gradients satisfy
\begin{equation}
\begin{aligned}
\frac{\partial\mathcal L_{\mathrm{post}}}{\partial P_j}
&=P_j-R_j,\\
\frac{\partial\mathcal L_{\mathrm{pre}}}{\partial P_j}
&=\sum_{i=1}^{N}A_{ij}\bigl((AP)_i-R_i\bigr).
\end{aligned}
\label{eq:locality-gradients}
\end{equation}
\end{theorem}

\begin{proof}
The $i$th rows are $(AX_q)_i+P_i$ and $(AX_q)_i+\sum_j A_{ij}P_j$. Differentiation gives the Jacobian blocks. Applying the chain rule to the rowwise squared reconstruction residuals gives the gradients; Appendix~A.2 provides the full derivation.
\end{proof}

At the insertion site, each post-aggregation prompt vector adjusts only its corresponding receiver, and its reconstruction gradient depends solely on that receiver's corrected residual. By contrast, a pre-aggregation prompt affects every receiver with $A_{ij}\ne0$, including its own node when $A_{jj}\ne0$.

\subsection{Requantization and prediction stability}
LoRAP's corrected representation $Z_P=\widehat S+P$ is requantized for subsequent low-precision operations. Fix the post-insertion quantizer pair $Q,DQ$, with nearest rounding and scale $\Delta>0$, and assume no clipping occurs for the inputs under consideration. In real units, define
\begin{equation}
\begin{aligned}
\widetilde S(P)&=DQ(Q(Z_P)),\\
\xi_q(P)&=\widetilde S(P)-Z_P,\\
e_q(P)&=\|S_{\mathrm{fp}}-\widetilde S(P)\|_F.
\end{aligned}
\label{eq:requantized-error}
\end{equation}

\begin{theorem}[Requantization error bound]
\label{thm:requantization}
Under the assumptions above, the actual residual and rounding error satisfy
\begin{equation}
\begin{aligned}
S_{\mathrm{fp}}-\widetilde S(P)&=R-P-\xi_q(P),\\
\|\xi_q(P)\|_F&\le b_q:=\tfrac{\Delta}{2}\sqrt{Nd}.
\end{aligned}
\label{eq:requantized-residual}
\end{equation}
Consequently,
\begin{equation}
|e_q(P)-\varepsilon(P)|\le b_q,\qquad
e_q(P)\le\varepsilon(P)+b_q.
\label{eq:requantization-bound}
\end{equation}
If $P=R$ and $S_{\mathrm{fp}}$ is within the representable interval, only rounding error remains. If its entries also lie on the quantization grid, $e_q(P)=0$.
\end{theorem}

\begin{proof}
Substituting $\widetilde S(P)=Z_P+\xi_q(P)$ gives the residual identity. Each entry incurs rounding error at most $\Delta/2$, giving the Frobenius bound. The triangle and reverse triangle inequalities yield the two error bounds. See Appendix~A.3 for the full proof and extensions.
\end{proof}

For the same quantizer, let $e_{\mathrm{base}}=\|S_{\mathrm{fp}}-DQ(Q(\widehat S))\|_F$ be the zero-prompt error. The correction benefit survives requantization whenever
\begin{equation}
\varepsilon(P)+b_q<e_{\mathrm{base}}
\quad\Longrightarrow\quad e_q(P)<e_{\mathrm{base}}.
\label{eq:correction-benefit}
\end{equation}
Thus any realized prompt, including LoRAP's low-rank output, reduces the post-requantization error whenever its continuous residual $\varepsilon(P)$ satisfies \eqref{eq:correction-benefit}.

Appendix~A.3 extends this single-layer bound to an $L$-layer GNN with full-precision outputs $H_\ell$ and prompted quantized outputs $\widetilde H_\ell$. When each reference layer is $L_\ell$-Lipschitz in the Frobenius norm and both branches start from the same input, the final representation error $e_L:=\|\widetilde H_L-H_L\|_F$ satisfies
\begin{equation}
e_L\le\sum_{\ell=1}^{L}
\left(\prod_{j=\ell+1}^{L}L_j\right)\delta_\ell,
\label{eq:propagated-error}
\end{equation}
where $\delta_\ell$ is the Frobenius discrepancy between the $\ell$th full-precision and prompted quantized layer outputs evaluated at $\widetilde H_{\ell-1}$ (accounting for aggregation, weight update, and activation errors). When $H_L$ and $\widetilde H_L$ are the reference and prompted quantized logit matrices with node rows $z_i$ and $\widetilde z_i$, suppose node $i$ has a unique reference prediction $y_i^*=\argmax_c z_{i,c}$ with margin $m_i:=z_{i,y_i^*}-\max_{c\ne y_i^*}z_{i,c}>0$. Since $\|\widetilde z_i-z_i\|_\infty\le e_L$, having $e_L<m_i/2$ (or more generally $\|\widetilde z_i-z_i\|_\infty<m_i/2$) preserves the predicted class $y_i^*$ without requiring zero reconstruction error.
\end{purplerevisionblock}

\begin{lowrankrevisionblock}
\subsection{Transfer to low-rank prompt parameterization}
Appendix~A.6 extends Theorems~\ref{thm:optimization-geometry}--\ref{thm:requantization} from unconstrained prompt outputs $P\in\mathbb{R}^{N\times d}$ to LoRAP's shared low-rank Softmax parameterization $P(\theta_P)=\mathrm{Softmax}(\phi(\widehat S))P_A P_B$ introduced in Section~V. In factor coordinates, post-aggregation curvature remains independent of $A$, and for fixed routing weights every critical point of $(P_A,P_B)$ is either a global minimum or a strict saddle with no spurious local minima, whereas pre-aggregation low-rank prompting inherits spectral scaling and flat directions from $A^\top A$. For shared parameters, post-aggregation gradients pair each node's feature and routing weight solely with its own unmixed receiver residual rather than the neighbor-mixed residual induced by $A$.
\begin{secondlowrankrevisionblock}%
Moreover, when $k\ge r+1$ ($r<k$), Softmax convex combinations can span both signs across the top-$r$ singular subspace of $R$, extending the post-requantization guarantee~\eqref{eq:correction-benefit} to both uniform and heterogeneous quantization scales.%
\end{secondlowrankrevisionblock}
\end{lowrankrevisionblock}
\begin{figure*}[!t]
\centering
\includegraphics[width=0.6\linewidth]{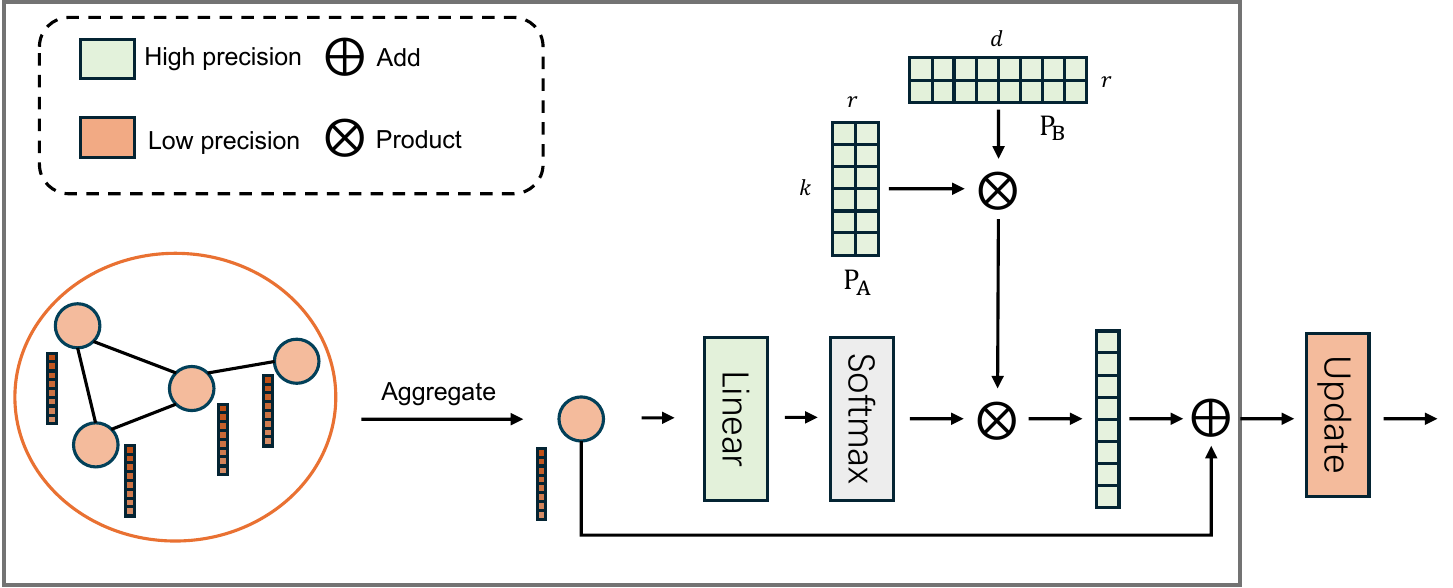}
\caption{To insert low-rank prompts in an aggregation, we first get the aggregated feature for a set of quantized nodes.
Then the aggregated feature is used to perform a weighted sum of the basis vectors (composed of low-rank matrices) to produce the prompt
The prompt is then added to the aggregated feature. The prompt generation procedure is in high-precision, other GNN operations are in low-precision.}
\label{fig:motivation3}
\end{figure*}
\section{Low-Rank Aggregation Prompt Tuning}
Fig.~\ref{fig:motivation3} illustrates the proposed method, whose components are discussed in the following subsections.
\begin{purplerevisionblock}
\subsection{Aggregation prompt insertion}
At layer $l$, LoRAP inserts node-dependent prompt vectors into the dequantized aggregation output $\widehat{\mathbf{S}}^{(l)}\in\mathbb{R}^{N\times d}$ (corresponding to $\widehat S$ in Section~\ref{sec:prompt-theory}). Let $\widehat{\mathbf{s}}_i^{(l)}$ denote the $i$th row of $\widehat{\mathbf{S}}^{(l)}$, and stack the node prompt vectors $\mathbf{p}_i^{(l)}\in\mathbb{R}^{1\times d}$ into $\mathbf{P}_s^{(l)}\in\mathbb{R}^{N\times d}$ (denoted by $P$ in Section~\ref{sec:prompt-theory}). Following \eqref{eq:prompt-residual}, the continuous corrected output is
\begin{equation}
\mathbf{Z}_P^{(l)}=\widehat{\mathbf{S}}^{(l)}+\mathbf{P}_s^{(l)},\qquad
\bigl(\mathbf{Z}_P^{(l)}\bigr)_i=\widehat{\mathbf{s}}_i^{(l)}+\mathbf{p}_i^{(l)}.
\label{AggrPrompt}
\end{equation}
As established by Theorems~\ref{thm:optimization-geometry} and~\ref{thm:insertion-locality}, this placement gives each prompt direct receiver-level access to the aggregation residual $\mathbf{R}^{(l)}=\mathbf{S}_{\mathrm{fp}}^{(l)}-\widehat{\mathbf{S}}^{(l)}$ under the reconstruction metric \eqref{eq:prompt-objectives}, while Theorem~\ref{thm:requantization} characterizes how the achieved residual $\varepsilon(\mathbf{P}_s^{(l)})$ carries through requantization.
\end{purplerevisionblock}
\subsection{Input-dependent low-rank aggregation prompt}
\begin{purplerevisionblock}
Different neighborhoods can require different corrections. To generate node-dependent prompts without storing a separate trainable vector for every node, LoRAP combines a small set of shared prompt bases using input-dependent weights.
Specifically, for each layer $l$ in a GNN with $L$ layers, we learn a set of $k$ prompt bases $\mathcal{B}^{(l)}=\{\mathbf{b}^{(l)}_1, \mathbf{b}^{(l)}_2, \cdots, \mathbf{b}^{(l)}_{k}\}$, where each $\mathbf{b}^{(l)}_{m} \in \mathbb{R}^{d}$ and $k$ is a hyperparameter.
Using $k\ll N$ reduces the number of trainable basis vectors.
We then propose to use a high-precision input-dependent linear function $\phi$ followed by a softmax to map the dequantized aggregated feature $\widehat{\mathbf{s}}_i^{(l)}$ to a probability distribution \(\boldsymbol{\alpha}_i^{(l)} \in \mathbb{R}^k\) where the element \(\alpha_{im}^{(l)}\) represents the weight or importance of the corresponding basis prompt vector \(\mathbf{b}^{(l)}_{m}\), \textit{i.e.},
\begin{equation}
    \boldsymbol{\alpha}_i^{(l)} = \mathrm{Softmax}\bigl(\phi(\widehat{\mathbf{s}}_i^{(l)})\bigr) \,.
\end{equation}

\end{purplerevisionblock}
The probabilities \(\boldsymbol{\alpha}_i^{(l)}\) are positive and ensure that the weights sum to 1, which means they can be interpreted as the relative importance of each basis vector in constructing the final output.
\purplerevision{In our implementation, layers sharing the same aggregation width $d$ share the linear projection $\phi$ while maintaining layer-specific prompt bases $\mathcal{B}^{(l)}$ (layers with different aggregation widths use layer-specific projections).}
Using these learned probabilities, the prompt \(\mathbf{p}^{(l)}_{i} \in \mathbb{R}^d\) is defined as the convex combination of basis vectors
\begin{equation}
    \mathbf{p}^{(l)}_{i} = \sum_{m=1}^k \alpha_{im}^{(l)}\,\mathbf{b}^{(l)}_{m}\,.
\end{equation}
The generated prompt is input-dependent as it is dynamically formed by combining the learned basis vectors with weights that are produced based on the aggregation features.
\begin{table*}[t]
    \tablefontsize
 
    \centering
    \caption{ 
    The results of 3 quantization frameworks with FP32 baselines at the top. 
    We provide three quantization frameworks for INT4: standard QAT, Degree-Quant (DQ), and $A^2Q$.
    They are validated on 3 model architectures with \revision{four adaptation settings: (1) no prompts, (2) GPF-plus, (3) GraphLoRA, and (4) GPF-LoRAP}.
    For each quantization framework and model architecture, the best result among the \revision{four settings} is highlighted in bold\revision{, while the second-best result is shown in italics. Rankings use unrounded means.}
    }
    \begin{tabular}{p{0.7cm} c c c c c c c}
        \toprule
        Quant. & Model & \revision{Adaptation} &\multicolumn{2}{c}{\tablefontsizeXS Node Classification (Accuracy \%)} & \multicolumn{2}{c}{\tablefontsizeXS Graph Classification (Accuracy \%)} & \multicolumn{1}{c}{\tablefontsizeXS Graph Regression (Loss)} \\
        
        Frame. & Arch. & Method & Cora $\uparrow$ & Citeseer $\uparrow$ & MNIST $\uparrow$ & CIFAR-10 $\uparrow$ & ZINC $\downarrow$ \\
        \midrule
        
        \multirow{3}{*}{\parbox{0.7cm}{FP32}} 
        & GIN & \multirow{3}{*}{None} 
        & $77.6\pm1.1$ 
        & $66.1\pm0.9$ 
        & $96.4\pm0.4$ 
        & $53.3\pm3.7$ 
        & $0.414\pm0.009$ \\
        
        & GCN & 
        & $81.4\pm0.7$ 
        & $71.1\pm0.7$ 
        & $90.9\pm0.4$ 
        & $54.5\pm0.1$ 
        & $0.469\pm0.002$ \\
        
        & GAT & 
        & $83.1\pm0.4$ 
        & $72.5\pm0.7$ 
        & $95.8\pm0.4$ 
        & $65.4\pm0.4$ 
        & $0.463\pm0.002$ \\
        
        \midrule
        \midrule
        
        \multirow{4}{*}{\parbox{0.7cm}{QAT}}
        & \multirow{4}{*}{GIN}
        & None 
        & $44.2\pm1.5$ 
        & $18.7\pm3.7$ 
        & $91.0\pm0.5$ 
        & $48.7\pm1.7$ 
        & $\mathit{0.555\pm0.024}$ \\
        
        & & GPF-plus 
        & $\bm{45.6\pm3.0}$ 
        & $21.6\pm2.9$
        & $91.3\pm0.3$
        & $\mathit{55.3\pm0.9}$
        & $0.569\pm0.020$ \\

        & & \revision{GraphLoRA}
        & \revision{$37.0\pm1.7$}
        & \revision{$\mathit{25.7\pm4.5}$}
        & \revision{$\mathit{95.6\pm0.7}$}
        & \revision{$51.8\pm4.2$}
        & \revision{$0.589\pm0.010$} \\
        
        & & GPF-LoRAP 
        & $\mathit{45.1\pm2.8}$
        & $\revision{\bm{28.6\pm2.4}}$
        & $\revision{\bm{96.6\pm0.2}}$
        & $\bm{55.8\pm0.5}$ 
        & $\bm{0.528\pm0.017}$ \\
        
        \midrule
        
        \multirow{4}{*}{\parbox{0.7cm}{DQ}}
        & \multirow{4}{*}{GIN}
        & None 
        & $61.0\pm6.9$ 
        & $58.4\pm2.1$ 
        & $92.4\pm0.4$
        & $50.7\pm1.6$ 
        & $0.540\pm0.019$ \\
        
        & & GPF-plus 
        & $64.3\pm5.0$
        & $\mathit{63.1\pm3.5}$
        & $91.9\pm0.3$ 
        & $51.3\pm1.2$
        & $0.555\pm0.018$ \\

        & & \revision{GraphLoRA}
        & \revision{$\mathit{72.8\pm1.8}$}
        & \revision{$58.5\pm3.4$}
        & \revision{$\mathit{93.1\pm0.7}$}
        & \revision{$\mathit{51.7\pm0.8}$}
        & \revision{$\mathit{0.539\pm0.007}$} \\
        
        & & GPF-LoRAP 
        & $\revision{\bm{75.2\pm0.6}}$
        & $\bm{64.6\pm2.3}$ 
        & $\revision{\bm{94.3\pm0.3}}$
        & $\bm{53.0\pm0.9}$ 
        & $\bm{0.512\pm0.008}$ \\
        
        \midrule
        
        \multirow{4}{*}{\parbox{0.7cm}{$A^2Q$}}
        & \multirow{4}{*}{GIN}
        & None 
        & $\mathit{77.4\pm0.9}$
        & $\mathit{66.6\pm0.1}$
        & $95.7\pm0.2$ 
        & $54.9\pm0.9$ 
        & $0.400\pm0.016$ \\
        
        & & GPF-plus 
        & $72.7\pm0.8$ 
        & $59.5\pm6.6$ 
        & $\mathit{95.9\pm0.4}$
        & $\mathit{56.8\pm1.4}$
        & $\mathit{0.399\pm0.009}$ \\

        & & \revision{GraphLoRA}
        & \revision{$66.8\pm1.7$}
        & \revision{$38.3\pm2.9$}
        & \revision{$95.4\pm0.1$}
        & \revision{$54.1\pm1.4$}
        & \revision{$0.462\pm0.009$} \\
        
        & & GPF-LoRAP 
        & $\bm{78.5\pm1.0}$ 
        & $\bm{70.7\pm0.6}$ 
        & $\bm{96.4\pm0.2}$ 
        & $\bm{57.4\pm1.4}$ 
        & $\bm{0.361\pm0.010}$ \\
        
        \midrule
        \midrule
        
        \multirow{4}{*}{\parbox{0.7cm}{QAT}}
        & \multirow{4}{*}{GCN}
        & None 
        & $66.4\pm7.0$ 
        & $\mathit{63.4\pm2.2}$
        & $70.6\pm2.4$ 
        & $38.5\pm1.3$ 
        & $\mathit{0.707\pm0.056}$ \\
        
        & & GPF-plus 
        & $69.8\pm6.5$
        & $61.6\pm3.1$ 
        & $\mathit{71.5\pm1.3}$
        & $41.0\pm1.2$
        & $0.733\pm0.018$ \\

        & & \revision{GraphLoRA}
        & \revision{$\mathit{70.1\pm3.9}$}
        & \revision{$61.5\pm4.8$}
        & \revision{$40.3\pm1.7$}
        & \revision{$\mathit{43.6\pm0.9}$}
        & \revision{$1.221\pm0.043$} \\
        
        & & GPF-LoRAP 
        & $\bm{71.3\pm8.3}$ 
        & $\bm{63.9\pm2.2}$ 
        & $\bm{73.6\pm1.2}$ 
        & $\revision{\bm{47.2\pm2.2}}$
        & $\bm{0.690\pm0.021}$ \\
        
        \midrule
        
        \multirow{4}{*}{\parbox{0.7cm}{DQ}}
        & \multirow{4}{*}{GCN}
        & None 
        & $\mathit{72.5\pm3.7}$
        & $\mathit{68.6\pm3.3}$ 
        & $87.0\pm1.2$ 
        & $49.5\pm0.8$ 
        & $0.657\pm0.032$ \\
        
        & & GPF-plus 
        & $72.3\pm2.8$ 
        & $\bm{71.4\pm3.1}$ 
        & $\mathit{90.9\pm0.7}$
        & $\mathit{51.1\pm0.5}$
        & $\mathit{0.534\pm0.017}$ \\

        & & \revision{GraphLoRA}
        & \revision{$65.1\pm9.7$}
        & \revision{$52.1\pm4.7$}
        & \revision{$87.5\pm1.1$}
        & \revision{$44.9\pm1.0$}
        & \revision{$0.568\pm0.010$} \\
        
        & & GPF-LoRAP 
        & $\bm{78.0\pm2.5}$ 
        & $66.1\pm1.3$
        & $\bm{92.4\pm0.5}$ 
        & $\bm{53.4\pm0.5}$ 
        & $\revision{\bm{0.529\pm0.014}}$ \\
        
        \midrule
        
        \multirow{4}{*}{\parbox{0.7cm}{$A^2Q$}}
        & \multirow{4}{*}{GCN}
        & None 
        & $76.1\pm0.3$ 
        & $69.7\pm1.6$ 
        & $90.5\pm0.5$ 
        & $52.4\pm0.8$ 
        & $\mathit{0.504\pm0.009}$ \\
        
        & & GPF-plus 
        & $\mathit{78.0\pm0.1}$
        & $\mathit{70.0\pm2.3}$
        & $\mathit{91.0\pm0.5}$
        & $\mathit{53.2\pm1.0}$
        & $0.506\pm0.008$ \\

        & & \revision{GraphLoRA}
        & \revision{$51.5\pm6.7$}
        & \revision{$38.5\pm7.5$}
        & \revision{$89.9\pm0.2$}
        & \revision{$49.9\pm0.5$}
        & \revision{$0.521\pm0.008$} \\
        
        & & GPF-LoRAP 
        & $\bm{79.1\pm0.1}$ 
        & $\bm{71.8\pm0.7}$ 
        & $\bm{91.8\pm0.6}$ 
        & $\bm{53.5\pm1.2}$ 
        & $\bm{0.482\pm0.011}$ \\
        
        \midrule
        \midrule
        
        \multirow{4}{*}{\parbox{0.7cm}{QAT}}
        & \multirow{4}{*}{GAT}
        & None 
        & $52.8\pm0.7$ 
        & $\mathit{30.1\pm14.5}$
        & $76.3\pm1.2$ 
        & $41.9\pm1.1$ 
        & $0.702\pm0.011$ \\
        
        & & GPF-plus 
        & $54.9\pm5.3$
        & $19.7\pm3.0$ 
        & $\mathit{77.5\pm0.9}$
        & $\mathit{43.1\pm1.2}$
        & $0.693\pm0.025$ \\

        & & \revision{GraphLoRA}
        & \revision{$\mathit{65.4\pm2.0}$}
        & \revision{$20.4\pm7.6$}
        & \revision{$74.6\pm2.1$}
        & \revision{$22.4\pm1.2$}
        & \revision{$\mathit{0.606\pm0.010}$} \\
        
        & & GPF-LoRAP 
        & $\bm{66.8\pm1.6}$ 
        & $\bm{40.6\pm7.3}$ 
        & $\bm{78.2\pm1.1}$ 
        & $\bm{43.8\pm0.7}$ 
        & $\bm{0.581\pm0.013}$ \\
        
        \midrule
        
        \multirow{4}{*}{\parbox{0.7cm}{DQ}}
        & \multirow{4}{*}{GAT}
        & None 
        & $68.7\pm7.5$
        & $41.9\pm5.9$
        & $90.3\pm0.5$ 
        & $54.2\pm0.5$
        & $\mathit{0.556\pm0.022}$ \\
        
        & & GPF-plus 
        & $58.3\pm5.4$ 
        & $39.1\pm9.6$ 
        & $93.1\pm0.6$
        & $51.5\pm0.8$ 
        & $0.589\pm0.018$ \\

        & & \revision{GraphLoRA}
        & \revision{$\mathit{69.5\pm8.0}$}
        & \revision{$\mathit{63.2\pm3.3}$}
        & \revision{$\mathit{94.7\pm0.4}$}
        & \revision{$\mathit{58.6\pm1.0}$}
        & \revision{$0.583\pm0.016$} \\
        
        & & GPF-LoRAP 
        & $\bm{72.5\pm2.3}$ 
        & $\revision{\bm{63.2\pm1.0}}$
        & $\revision{\bm{94.9\pm0.2}}$
        & $\revision{\bm{61.5\pm0.3}}$
        & $\bm{0.508\pm0.023}$ \\
        
        \midrule
        
        \multirow{4}{*}{\parbox{0.7cm}{$A^2Q$}}
        & \multirow{4}{*}{GAT}
        & None 
        & $82.6\pm0.8$ 
        & $71.2\pm0.9$ 
        & $\mathit{92.5\pm0.6}$
        & $61.9\pm1.2$ 
        & $0.674\pm0.007$ \\
        
        & & GPF-plus 
        & $\mathit{83.4\pm0.5}$
        & $\mathit{71.8\pm0.2}$
        & $92.4\pm0.8$ 
        & $\mathit{62.3\pm1.4}$
        & $0.685\pm0.008$ \\

        & & \revision{GraphLoRA}
        & \revision{$79.3\pm2.4$}
        & \revision{$67.2\pm1.8$}
        & \revision{$60.7\pm16.7$}
        & \revision{$61.4\pm0.7$}
        & \revision{$\mathit{0.581\pm0.023}$} \\
        
        & & GPF-LoRAP 
        & $\bm{83.6\pm0.6}$ 
        & $\bm{72.5\pm0.8}$ 
        & $\bm{93.1\pm0.7}$ 
        & $\bm{62.9\pm1.0}$ 
        & $\revision{\bm{0.547\pm0.008}}$ \\
        
        \bottomrule
    \end{tabular}    
    \label{tab:results}
\end{table*}

\begin{purplerevisionblock}
In our implementation, the set of prompt bases $\mathcal{B}^{(l)}$ is stacked as a matrix $\mathbf{B}^{(l)} \in \mathbb{R}^{k\times d}$.
Storing $\mathbf{B}^{(l)}$ across $L$ layers requires $L\times k \times d$ basis parameters.
To improve parameter efficiency, we factorize $\mathbf{B}^{(l)}$ into two high-precision low-rank matrices, $\mathbf{P}^{(l)}_A \in \mathbb{R}^{k\times r}$ and $\mathbf{P}^{(l)}_B \in \mathbb{R}^{r\times d}$, where the rank budget $r \ll \min(k,d)$, giving $\mathbf{B}^{(l)} = \mathbf{P}^{(l)}_A \mathbf{P}^{(l)}_B$.
\end{purplerevisionblock}
This in turn reduces the number of parameters from $L\times k \times d$ to $L\times r \times(k +d)$.
Given the quantized matrix of aggregated features $\mathbf{S}^{(l)}_q$, we then incorporate our input-dependent low-rank prompts into the $l$-th GNN layer as
\begin{gather}
    \mathbf{\hat{S}}^{(l)} = DQ(\mathbf{S}^{(l)}_q) \\
    \mathbf{P}^{(l)}_s = \textrm{Softmax}\left(\phi( \mathbf{\hat{S}}^{(l)})\right)\mathbf{P}^{(l)}_A \mathbf{P}^{(l)}_B \\
    \mathbf{S}^{(l)*}_q = Q( \mathbf{\hat{S}}^{(l)} + \mathbf{P}^{(l)}_s) \,,
\end{gather}
\purplerevision{where $DQ$ is the \emph{dequantization operator} defined in Section~\ref{appendix:quantization}. Since $\mathbf{B}^{(l)}=\mathbf{P}^{(l)}_A\mathbf{P}^{(l)}_B$, the realized prompt $\mathbf{P}^{(l)}_s$ has rank at most $r$, and Theorem~\ref{thm:requantization} bounds the post-requantization error of $DQ(\mathbf{S}^{(l)*}_q)$.}
Finally, the obtained prompted aggregation feature $\mathbf{S}^{(l)*}_q$ is used for further low-precision operations.
By doing so, the gradient flows through the quantized weights to the full-precision prompts during the optimization process, allowing the learned prompt to be aware of the quantization effects.
This awareness allows the model, when making predictions, to adapt its behavior to compensate for the low-bit representation, potentially leading to improved performance.
\begin{purplerevisionblock}
\subsection{QAT frameworks \& LoRAP integration}
\label{sec:joint-qat}
LoRAP retains the task objective $L_D$ of the host QAT framework (see Section~\ref{sec:quantization-techniques}), jointly updating the prompt parameters $\theta_P$ (routing map and low-rank factors), the backbone weights $W$, and any trainable quantizer parameters $\theta_Q$ without freezing the backbone or adding an auxiliary reconstruction loss:
\[
\min_{\theta_Q,\theta_P,W} L_D.
\]
Although these coupled task-loss updates differ from fixed-residual gradient descent, Appendix~A.4 shows that at every iteration $t$, the current backbone and quantizer state define $S_{\mathrm{fp},t}$, $\widehat S_t$, and the realized prompt $P_t$ for which Theorem~\ref{thm:requantization} bounds the post-requantization error pointwise via $\varepsilon_t(P_t)$ and the current scale $\Delta_t$.
\end{purplerevisionblock}
\subsection{LoRAP kernel}

\begin{figure}[!t]
\centering
\includegraphics[width=0.9\linewidth]{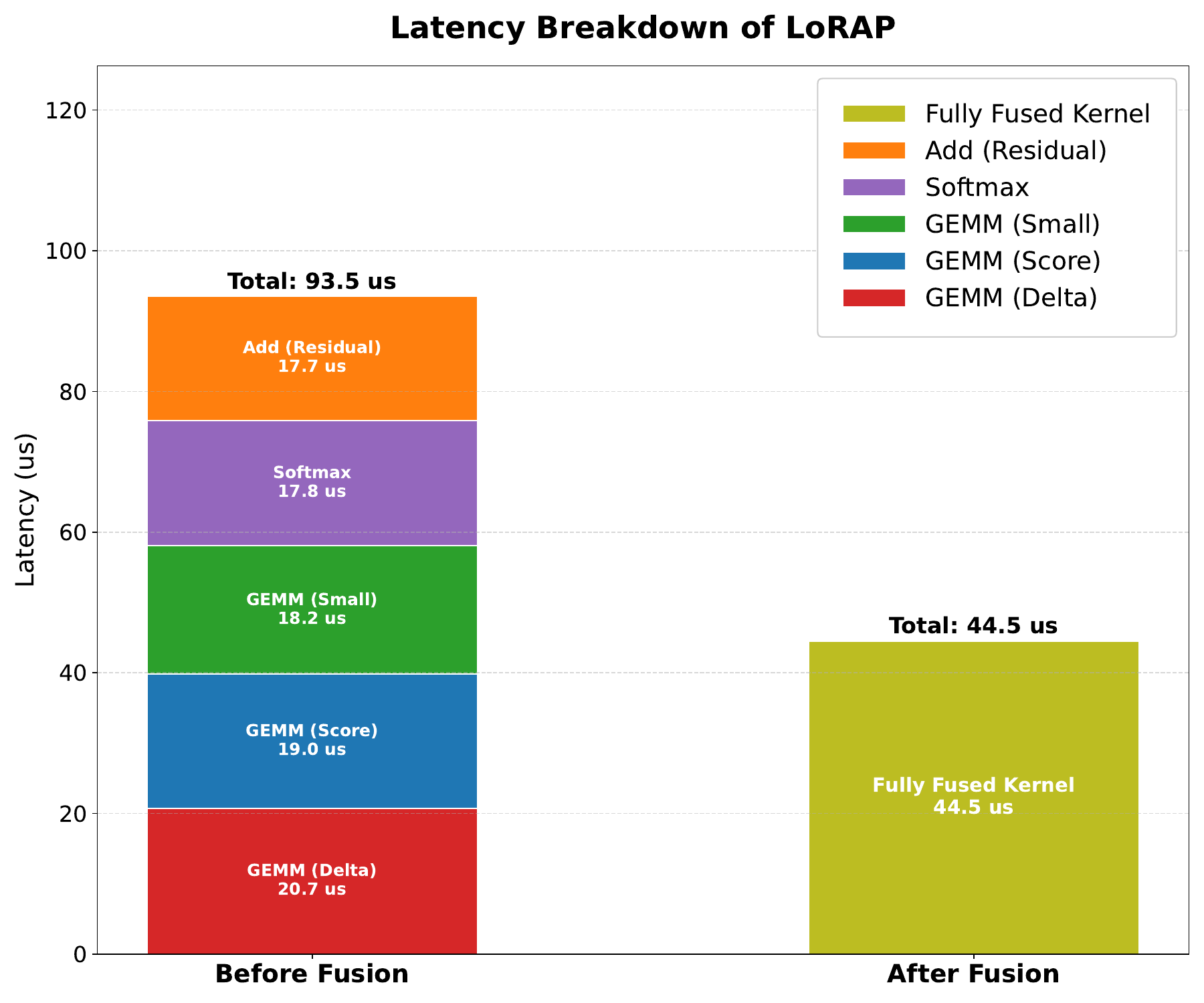}
\caption{Latency comparison.}
\label{fig:kernel}
\end{figure}

\purplerevision{Although the low-rank prompt generator uses small factor matrices,} executing it as a series of sequential operations incurs a notable latency overhead, totaling 93.5 $\mu$s as illustrated in Fig. \ref{fig:kernel}. For such lightweight projections (e.g., GEMMs) and element-wise additions, the bottleneck shifts from computation to memory access, as intermediate activations must be repeatedly loaded from and stored to DRAM. To address this, we propose to fuse the LoRAP kernel using Triton \cite{tillet2019triton}, a Python-based
GPU programming framework. By fusing the low-rank prompt generation and the residual addition directly into the aggregation operation, the LoRAP kernel allows intermediate states to be exchanged via the GPU register file or shared memory. This design eliminates redundant global memory accesses and kernel launch overheads, reducing the latency to 44.5 $\mu$s (a 2.1$\times$ speedup) and effectively rendering the prompt injection process more efficient.

\section{Experiments}
We integrate our prompt tuning strategy into 4 graph QAT frameworks to restore their performance under INT4 quantization bit-width, specifically standard QAT \cite{jacob2018quantization}, Degree-Quant (DQ) \cite{tailor2020degree}, $A^2Q$ \cite{zhu2023rm}, and MixQ \cite{moustafa2025efficient}. 
The GNN architectures employed in our experiments include GIN \cite{xu2018powerful}, GCN \cite{kipf2017semi}, and GAT \cite{velivckovic2017graph}.
Evaluation results are reported across both node-level and graph-level tasks. For node-level tasks, the Cora, CiteSeer\purplerevision{, and PubMed} datasets are utilized, as well as large-scale OGB benchmarks (ogb-arxiv, ogb-products, and ogbn-mag) to verify scalability.
For graph-level tasks, the REDDIT-BINARY, MNIST, CIFAR10, and ZINC datasets are adopted, where ZINC is specifically used for regression tasks.
To provide a comparative baseline, we employ SOTA graph prompting methods, including GPF-plus, UniPrompt \cite{huangone}, and EdgePrompt+ \cite{fuedge}\purplerevision{, as well as the low-rank graph adapter GraphLoRA \cite{graphlora2024} (adapted to QAT as detailed in Appendix~B.1),} for performance recovery of quantized GNNs.
Detailed descriptions of the datasets are provided in \purplerevision{Section~\ref{sec:datasets}}\footnote{Note that, to ensure a fair and correct comparison, we follow the same experimental setting (including datasets and model architectures) of the QAT frameworks our work is based on.}. 
Since GPF-plus and LoRAP can be seen as complementary approaches, our implementation integrates GPF-plus into the input feature space and LoRAP into the aggregation feature space, denoted as GPF-LoRAP.
The additional hyperparameters introduced in our approach include the number of prompts ($k$) and the rank ($r$).
For fair comparison, other hyperparameters retain the same configurations as those used in the original quantization frameworks, with only $k$ and $r$ subjected to tuning.
We also conduct experiments on full-precision (FP32) models to determine if prompt tuning can bring the performance of low-precision GNNs to that of FP32 ones.
We then evaluate inference latency to show that incorporating prompts into quantized GNNs introduces a negligible computational overhead.
Finally, we perform ablation studies to separate the effect of different prompting strategies.

\subsection{Datasets}
\label{sec:datasets}

\begin{table*}[ht!]
    \caption{Overall statistics for the datasets used in this work.}
    \begin{center}
    \begin{tabular}{ccccccc}
        \hline \toprule[2pt]
        Task                             & Name          & Graphs & Nodes       & Edges       & Features & Classes \\ \midrule[1pt]
        \multirow{6}{*}{Node-level}      & Cora          & 1      & 2,708       & 10,556      & 1,433    & 7       \\
                                         & CiteSeer      & 1      & 3,327       & 9,104       & 3,703    & 6       \\
                                         & \purplerevision{PubMed} & \purplerevision{1} & \purplerevision{19,717} & \purplerevision{88,648} & \purplerevision{500} & \purplerevision{3} \\
                                         & OGB-Arxiv     & 1      & 169,343     & 1,166,243   & 128      & 40      \\
                                         & OGB-Products  & 1      & 2,449,029   & 61,859,140  & 100      & 47      \\ 
                                         & OGB-Mag       & 1      & 1,939,743   & 25,582,108  & 128      & 349      \\ \hline
        \multirow{4}{*}{Graph-level}     & REDDIT-BINARY & 2000   & $\sim$429.6 & $\sim$995.5 & 0        & 2       \\
                                         & MNIST         & 70000  & $\sim$71    & $\sim$565   & 3        & 10      \\
                                         & CIFAR10       & 60000  & $\sim$117.6 & $\sim$941.2 & 5        & 10      \\
                                         & ZINC          & 12000  & $\sim$23    & $\sim$49.8  & 28       & ---     \\ \bottomrule[2pt]
    \end{tabular}
    \end{center}
    \label{dataset_sta}
\end{table*}

Table~\ref{dataset_sta} summarizes the key properties of all datasets considered in this paper. For node-level tasks, we utilize the standard citation networks Cora, CiteSeer\purplerevision{, and PubMed}, in which each node represents a document and each edge represents a citation link. To verify scalability, we also incorporate large-scale OGB benchmarks, including OGB-Arxiv, OGB-Products, and the heterogeneous OGB-Mag. In these datasets, node features are primarily bag-of-words vectors or pretrained features, and the task is to predict node classes in a semi-supervised setting \cite{kipf2017semi}.

For graph-level tasks, MNIST and CIFAR-10 are image-classification datasets converted into superpixel graphs using SLIC, where nodes correspond to perceptually similar pixel clusters \cite{6205760}. ZINC consists of molecular graphs with atoms as nodes and bonds as edges for regression tasks \cite{jin2018junction}. \purplerevision{REDDIT-BINARY is a social network graph dataset} extracted from Reddit discussion threads, where nodes represent users and edges denote interactions between users.

MNIST, CIFAR-10, ZINC, and OGB use their standard train/validation/test splits. Citation networks use the semi-supervised split from \cite{kipf2017semi}; \purplerevision{REDDIT-BINARY} performance is reported as mean accuracy over 10-fold cross-validation.

\subsection{Experimental Setup}
To ensure a fair comparison, we replicate the GNN architectures and FP32 baselines from DQ \cite{tailor2020degree} and A\textsuperscript{2}Q \cite{zhu2023rm} across all tasks. Table \ref{tab:arch-config} details the specific configurations (number of layers, hidden-unit sizes, and the presence of skip connections) used in our experiments. We implement the QAT, DQ, and A\textsuperscript{2}Q frameworks in PyTorch Geometric \cite{fey2019fast}, adhering to the original hyperparameter settings of each method. All quantized models are trained with the Adam optimizer; the only parameters we vary are the number of prompts \(k\), selected from \(\{5,10,20,40\}\), and the prompt rank \(r\), chosen from \(\{1,2,4,8\}\) \purplerevision{with $r<k$}.  
Results on Cora, CiteSeer, and ogb-arxiv are averaged over 100 independent runs with different random seeds, while results on MNIST, CIFAR-10, and ZINC are averaged over 10 runs. For REDDIT-BINARY, we perform 10-fold cross-validation with a fixed split seed of 12345.

For the classification tasks (Cora, CiteSeer\purplerevision{, PubMed}, MNIST, CIFAR-10, REDDIT-BINARY), we report the accuracy
\begin{equation}
\mathrm{Accuracy} = \frac{1}{N}\sum_{i=1}^N \mathbf{1}(y_i = \hat y_i),
\end{equation}
where \(y_i\) and \(\hat y_i\) are the true and predicted labels, respectively, and \(\mathbf{1}(\cdot)\) is the indicator function.

For the regression task on ZINC, we use the Mean Absolute Error (MAE) loss
\begin{equation}
\mathrm{MAE} = \frac{1}{N}\sum_{i=1}^N \bigl|y_i - \hat y_i\bigr|,
\end{equation}
where \(y_i\) is the ground-truth property value and \(\hat y_i\) is the model prediction.

As for the experimental environment, all experiments were carried out on servers featuring an AMD EPYC 9124 16-core CPU paired with four NVIDIA RTX 4090 GPUs and running Ubuntu 22.04. Each node has 512 GB of RAM, and we used CUDA 12.7 with PyTorch 2.7.0 for all GPU-accelerated workloads.

\begin{table*}[t!]
\caption{Architectural configurations on selected datasets.}
\centering
\begin{tabular}{llcccccc}
\toprule
Model & Property         & Cora          & CiteSeer      & MNIST       & CIFAR10     & ZINC        & REDDIT-BINARY \\
\midrule
\multirow{3}{*}{GCN} & Layers          & 2             & 2             & 5+1 MLP     & 5+1 MLP     & 5+1 MLP     & --            \\
                     & Hidden units    & 16      & 16      & 146         & 146         & 145         & --            \\
                     & Skip connection & NO            & NO            & YES         & YES         & YES         & --            \\
\midrule
\multirow{3}{*}{GIN} & Layers          & 2             & 2             & 5+1 MLP     & 5+1 MLP     & 5+1 MLP     & 5             \\
                     & Hidden units    & 16      & 16      & 110         & 110         & 110         & 64            \\
                     & Skip connection & NO            & NO            & YES         & YES         & YES         & NO            \\
\midrule
\multirow{3}{*}{GAT} & Layers          & 2             & 2             & 5+1 MLP     & 5+1 MLP     & 5+1 MLP     & --            \\
                     & Hidden units    & $8\times h$  & $8\times h$  & $19\times h$& $19\times h$& $18\times h$& --            \\
                     & Skip connection & NO            & NO            & YES         & YES         & YES         & --            \\
\bottomrule
\end{tabular}

\label{tab:arch-config}
\end{table*}

\section{Results}

\subsection{Main results}
We show how our prompt learning strategy improves the performance of 3 INT4 QAT frameworks in the Table~\ref{tab:results}. Full-precision (FP32) baselines are shown at the top. Under the naive QAT framework (INT4 QAT), all architectures suffer severe accuracy drops when no prompt is used (\emph{e.g.}, for GIN, Cora accuracy falls from $77.6\%$ to $44.2\%$).
Introducing GPF-plus yields only a marginal recovery, whereas GPF-LoRAP consistently narrows this gap, improving GIN by +0.9\%/+9.9\% on Cora/Citeseer, GCN by +4.9\%/+0.5\%, and GAT by +14.0\%/+10.5\%, while also boosting graph classification (+7.1\% for GIN on CIFAR-10).
Under DQ, which already retains more of the original accuracy, GPF-LoRAP further boosts performance across the board, confirming its generality.
Even with the $A^2Q$ framework, which already narrows the FP32 gap, GPF-LoRAP delivers consistent improvements.
\purplerevision{By contrast, while GraphLoRA is competitive in several DQ settings (\emph{e.g.}, reaching 72.8\% on Cora with GIN), it exhibits marked instability under $A^2Q$ and standard QAT (\emph{e.g.}, dropping from 66.6\% to 38.3\% on CiteSeer under $A^2Q$-GIN and from 70.6\% to 40.3\% on MNIST under QAT-GCN), indicating that weight-side adapters are less reliable under aggressive quantization.}
Remarkably, when combined with the $A^2Q$ quantization, GPF-LoRAP not only recovers nearly all of the performance lost to low-bit compression but actually surpasses the original FP32 baseline in several cases.
For example, on Cora, GIN quantized with $A^2Q$ and equipped with our prompts achieves 78.5\% accuracy, outperforming the 77.6\% of the full-precision mode.
In summary, GPF-LoRAP outperforms both the unprompted baseline and the GPF-plus strategy, often halving the remaining accuracy gap with the FP32 model.
The benefits are pronounced across nearly all settings, demonstrating that GPF-LoRAP can act as a general training strategy for GNN QAT frameworks.

\begin{table}[!htbp]
    \centering
    \caption{Results for QAT \& DQ GIN in INT4 \& 8 on Reddit-Binary. 
    }
    \setlength{\tabcolsep}{3pt}  
        \begin{tabular}{c c c c} 
            \toprule
            Quant. Frame. & Model & Prompt Method & \makecell[c]{REDDIT-BIN\\ (Acc.~\%) $\uparrow$ } \\
            \midrule
            FP32 & GIN & None & $92.2\pm2.3$ \\
            \midrule
            \multirow{3}{*}{QAT-W8A8} & \multirow{3}{*}{GIN} & None & $76.1\pm7.5$ \\
             &  & GPF-plus & $72.2\pm3.7$ \\
            &  & GPF-LoRAP& $81.2\pm2.7$ (+$\bm{5.1}$) \\
            \midrule
            \multirow{3}{*}{DQ-W8A8} & \multirow{3}{*}{GIN} & None & $91.8\pm2.3$ \\
             &  & GPF-plus & $94.1\pm2.0$ \\
             &  & GPF-LoRAP& $94.6\pm2.7$ (+$\bm{2.8}$) \\
            \midrule
            \multirow{3}{*}{QAT-W4A4} & \multirow{3}{*}{GIN} & None & $52.4\pm5.8$ \\
             &  & GPF-plus & $53.8\pm5.7$ \\
             &  & GPF-LoRAP& $69.6\pm5.6$ (+$\bm{17.2}$) \\
            \midrule
            \multirow{3}{*}{DQ-W4A4} & \multirow{3}{*}{GIN} & None & $89.4\pm2.9$ \\
             &  & GPF-plus & $92.3\pm2.4$ \\
             &  & GPF-LoRAP& $93.0\pm3.1$ (+$\bm{3.6}$) \\
            \bottomrule
        \end{tabular}
    \label{tab:resultsReddit}
\end{table}
\subsection{Results on different bit widths}

Table \ref{tab:resultsReddit} presents the Reddit-Binary classification accuracy of GIN under different bit-widths, using both standard QAT and DQ, with three prompting strategies: (1) no prompts, (2) node prompting with GPF-plus, and (3) GPF-LoRAP.

While standard GPF-plus is unstable and sometimes lowers the accuracy, GPF-LoRAP consistently improves performance (+5.1\% under QAT-W8A8) and even pushes the accuracy beyond the full-precision baseline in the stronger DQ-W8A8 setting (+2.8\%). Moreover, in the aggressive INT4 regime, when the model fails severely under QAT-W4A4, GPF-LoRAP provides a substantial recovery (+17.2\%).

\begin{table}[!htbp]
    \centering
        \caption{Ablation study for A2Q GIN on the MNIST dataset.
    }
     \setlength{\tabcolsep}{4pt}  
        \begin{tabular}{c c c c c } 
            \toprule
            Config & \makecell[c]{Average\\bits}  & \makecell[c]{\#Additional\\Parameters} &  \makecell[c]{MNIST\\(Acc.~\%)$\uparrow$}    \\
            \midrule
            FP32 & 32  & 0KB&$96.4\pm0.4$ \\
            \midrule
            INT4 & 4  &0KB& $94.4\pm0.2$ \\
            + GPF-plus & 4  &17.26KB& $96.0\pm0.5$ \\
            \revision{GraphLoRA} & \revision{4} & \revision{20.67KB} & \revision{$93.6\pm0.7$} \\
            \makecell[c]{+ GPF-plus + LoRAP\\(GPF-LoRAP)} & 4 &26.95KB & $96.2\pm0.4$ \\
            \midrule
            A2Q & 3.75 &0KB & $95.7\pm0.2$ \\
            + GPF-plus & 3.75 &17.26KB & $95.9\pm0.4$ \\
            \revision{GraphLoRA} & \revision{3.75} & \revision{20.67KB} & \revision{$95.4\pm0.1$} \\
            + EdgePrompt+ & 3.78 & 103.44KB & $95.5\pm0.3$\\
            + AP & 3.78  &43.03KB& $95.9\pm0.1$ \\
            + LoRAP & 3.76  &9.69KB & $96.2\pm0.3$ \\
            + GPF-plus + AP & 3.78 &60.29KB & $96.3\pm0.1$ \\
            \makecell[c]{+ GPF-plus + LoRAP\\(GPF-LoRAP)} & 3.76 &26.95KB  & $96.4\pm0.2$ \\
            
            \bottomrule
        \end{tabular}    
    \label{tab:resultsAblation}
\end{table}

{
\subsection{Results on large-scale datasets}
To verify the scalability of our approach, we extended our evaluation to three large-scale benchmarks: ogb-arxiv, ogb-products, and the heterogeneous graph ogbn-mag. As shown in Table \ref{tab:ogb_results_extended}, LoRAP consistently yields the highest accuracy across all tasks and quantization frameworks. A crucial observation is that standard node-level prompting method GPF-Plus struggle in these large-scale quantized settings. For instance, on the heterogeneous ogbn-mag dataset under the A2Q framework, applying GPF-Plus actually degrades the baseline performance from 32.7\% to 30.4\%. \purplerevision{In contrast, LoRAP raises the accuracy to 34.2\%.} Similarly, on ogb-products, LoRAP improves the MixQ baseline by a solid margin (+2.5\%), validating its robustness and effectiveness in handling diverse graph structures and scales.
}
\begin{table}[!htbp]
\centering
\caption{Performance comparison on large-scale datasets.}
\label{tab:ogb_results_extended}

\begin{tabular}{llcc}
\toprule
\textbf{Task} & \textbf{Method} & \textbf{Acc} & \textbf{Bits} \\
\midrule

\multirow{9}{*}{\shortstack[l]{ogb-\\arxiv\\(GCN)}} 
 & FP32 & $71.7 \pm 0.3$ & 32 \\
\cmidrule{2-4}
 & A2Q & $71.1 \pm 0.3$ & 2.7 \\
 & +GPF-Plus & $70.9 \pm 0.3$ & 2.7 \\
 & +UniPrompt & $71.3 \pm 0.4$ & 2.7 \\
 & +LoRAP & $73.6 \pm 0.1$ & 2.7 \\
\cmidrule{2-4}
 & MixQ & $69.3 \pm 0.0$ & 7.1 \\
 & +GPF-Plus & $70.1 \pm 0.2$ & 7.1 \\
 & +UniPrompt & $69.4 \pm 0.2$ & 7.1 \\
 & +LoRAP & $71.9 \pm 0.1$ & 7.0 \\

\midrule

\multirow{5}{*}{\shortstack[l]{ogb-products\\(GraphSAGE)}} 
 & FP32 & $66.6 \pm 1.3$ & 32 \\
\cmidrule{2-4}
 & MixQ & $60.8 \pm 2.2$ & 5.0 \\
 & +GPF-Plus & $61.0 \pm 1.6$ & 5.0 \\
 & +UniPrompt & $60.1 \pm 1.1$ & 4.9 \\
 & +LoRAP & $63.3 \pm 2.0$ & 5.0 \\

\midrule

\multirow{5}{*}{\shortstack[l]{ogbn-mag\\(GCN)}} 
 & FP32 & $33.6 \pm 0.4$ & 32 \\
\cmidrule{2-4}
 & A2Q & $32.7 \pm 0.4$ & 2.7 \\
 & +GPF-Plus & $30.4 \pm 0.1$ & 2.9 \\
 & +UniPrompt & $31.0 \pm 0.4$ & 2.9 \\
 & +LoRAP & $34.2 \pm 0.2$ & 2.7 \\

\bottomrule
\end{tabular}
\end{table}

\subsection{Latency comparison}

In Table \ref{tab:latency_speedup}, we report the latency for a single GCN layer on the CIFAR-10 dataset.
The hardware for testing is an AMD EPYC 9124 16-Core CPU paired with an Nvidia RTX4090 GPU. 

On the CPU, quantizing both weights and activations to 8 bits achieves a 2.39$\times$ speedup over the FP32 baseline. Incorporating the FP32 LoRAP prompt module on top of the W8A8 layer incurs minimal cost, still running almost 2.00$\times$ faster than the original FP32 layer. 
On the GPU, the efficiency is further improved by our fused kernel design. Specifically, the INT4+LoRAP (fused kernel) maintains a 1.86$\times$ speedup compared to the FP32 baseline, demonstrating that the overhead of prompt injection is effectively mitigated. \purplerevision{End-to-end full-graph GPU latency, peak memory, and model storage results on ogbn-arxiv are provided in Appendix~B.3.}

\begin{table}[!htbp]
\centering
\caption{Latency and Speedup comparison on CPU and GPU devices. Speedup is calculated relative to the FP32 baseline.}
\label{tab:latency_speedup}
\begin{tabular}{llcc}
\toprule
\textbf{Device} & \textbf{Setting} & \textbf{Latency} & \textbf{Speedup} \\
\midrule
\multirow{5}{*}{CPU} 
 & FP32 & 36.66ms & 1.00$\times$ \\
 & INT8 & 15.36ms & 2.39$\times$ \\
 & INT8+LoRAP & 18.29ms & 2.00$\times$ \\
 & INT4 & 11.78ms & 3.11$\times$ \\
 & INT4+LoRAP & 14.77ms & 2.48$\times$ \\
\midrule
\multirow{7}{*}{GPU} 
 & FP32 & 0.69ms & 1.00$\times$ \\
 & INT8 & 0.43ms & 1.60$\times$ \\
 & INT8+LoRAP & 0.52ms & 1.33$\times$ \\
 & INT8+LoRAP (fused) & 0.48ms & 1.44$\times$ \\
 & INT4 & 0.33ms & 2.09$\times$ \\
 & INT4+LoRAP  & 0.42ms & 1.64$\times$ \\
 & INT4+LoRAP (fused) & 0.37ms & 1.86$\times$ \\
\bottomrule
\end{tabular}
\end{table}

\begin{table}[!htbp]
\centering
\caption{GPF-LoRAP with {A\textsuperscript{2}Q} (INT4) for GIN on ZINC, mean $\pm$ std deviation for varying number of prompts $k$ and rank $r$.}
\setlength{\tabcolsep}{3.5pt}
\begin{tabular}{c|cccc}
\toprule
$r \backslash k$ & 5 & 10 & 20 & 40 \\
\midrule
1 & $0.382 \pm 0.013$ & $0.379 \pm 0.011$ & $0.378 \pm 0.009$ & $0.368 \pm 0.012$ \\
2 & $0.370 \pm 0.011$ & $0.372 \pm 0.005$ & $0.372 \pm 0.009$ & $0.361 \pm 0.010$ \\
4 & $0.378 \pm 0.014$ & $0.376 \pm 0.017$ & $0.373 \pm 0.011$ & $0.380 \pm 0.014$ \\
8 & \purplerevision{---} & $0.374 \pm 0.015$ & $0.374 \pm 0.015$ & $0.380 \pm 0.011$ \\
\bottomrule
\end{tabular}
\label{tab:zinc_results_adjusted}
\end{table}
\subsection{Ablation study}

Table~\ref{tab:resultsAblation} presents an ablation study on the MNIST dataset using $A^2Q$-quantized GIN models to isolate the impact of different adaptation strategies, including GPF-plus, \purplerevision{GraphLoRA,} EdgePrompt+, full-rank aggregation prompting (AP), and our proposed GPF-LoRAP.

EdgePrompt+ \purplerevision{and GraphLoRA both degrade} performance despite incurring higher parameter overhead \purplerevision{(103.44~KB and 20.67~KB, respectively)}. In contrast, the low-rank constraint in LoRAP demonstrates superior parameter efficiency by outperforming the full-rank AP strategy while requiring significantly fewer trainable parameters \purplerevision{(9.69~KB)}. Furthermore, there is a clear synergy between node and aggregation prompting: while GPF-plus and LoRAP individually improve over the baseline, their combination (GPF-LoRAP) yields the strongest gain, successfully bridging the gap to match the FP32 benchmark. \purplerevision{Additional control comparisons against an equal-capacity MLP and FP32 prompting on Cora, CiteSeer, and PubMed are reported in Appendix~B.2.}

\subsection{Impact of the Number of Prompts \texorpdfstring{\(k\)}{k} and the Aggregation Rank \texorpdfstring{\(r\)}{r}}

Table~\ref{tab:zinc_results_adjusted} reports the mean loss of GPF-LoRAP using {A\textsuperscript{2}Q} (INT4) and GIN on the ZINC dataset for different numbers of prompts \(k\) and aggregation rank \(r\) \purplerevision{with $r<k$}. For low ranks, as \(k\) increases from 5 to 40, the loss generally decreases: at \(r=1\) it drops from 0.382 to 0.368 and at \(r=2\) from 0.370 to the overall minimum of 0.361. For higher ranks (\(r=4\) and \(r=8\)), the lowest losses occur at \(k=20\) (0.373) \purplerevision{and \(k\in\{10,20\}\) (0.374), respectively,} before slightly rising at \(k=40\). Overall, the best performance is achieved at \(r=2, k=40\) with a mean loss of 0.361, confirming that moderate prompt counts and rank values yield the lowest prediction error. \purplerevision{The} optimal trade-off between prompt count and rank depends on the performance and computational budget.

\begin{table*}[!t]
\centering
\caption{Comparison of accuracy and bit-width on node and graph classification tasks.}
\label{tab:combined_results_split}

\begin{tabular}{l llcc llcc}
\toprule
\multirow{2}{*}{\textbf{Prompt Method}}
& \multicolumn{4}{c}{\textbf{Node Classification}}
& \multicolumn{4}{c}{\textbf{Graph Classification}} \\
\cmidrule(lr){2-5} \cmidrule(lr){6-9}
& \textbf{Quant.} & \textbf{Task} & \textbf{Acc} & \textbf{Bits}
& \textbf{Quant.} & \textbf{Task} & \textbf{Acc} & \textbf{Bits} \\
\midrule

None
& \multirow{6}{*}{DQ}
& \multirow{6}{*}{Cora-GCN}
& $72.5 \pm 3.7$ & 4.0
& \multirow{6}{*}{QAT}
& \multirow{6}{*}{Reddit-Binary}
& $52.4 \pm 5.8$ & 4.0 \\

GPF-Plus
& & & $72.3 \pm 2.8$ & 4.0
& & & $53.8 \pm 5.7$ & 4.0 \\

UniPrompt
& & & $72.8 \pm 2.4$ & 4.0
& & & - & - \\

EdgePrompt+
& & & $73.1 \pm 2.5$ & 4.0
& & & $53.2 \pm 4.9$ & 4.0 \\

LoRAP
& & & $77.7 \pm 1.8$ & 4.0
& & & $70.1 \pm 5.6$ & 4.0 \\

GPF-LoRAP
& & & $78.0 \pm 2.5$ & 4.0
& & & $69.6 \pm 5.6$ & 4.0 \\
\midrule

None
& \multirow{6}{*}{A2Q}
& \multirow{6}{*}{Cora-GCN}
& $76.1 \pm 0.3$ & 1.7
& \multirow{6}{*}{A2Q}
& \multirow{6}{*}{MNIST}
& $95.7 \pm 0.2$ & 3.8 \\

GPF-Plus
& & & $78.0 \pm 0.1$ & 1.7
& & & $95.9 \pm 0.4$ & 3.8 \\

UniPrompt
& & & $78.5 \pm 0.2$ & 1.7
& & & - & - \\

EdgePrompt+
& & & $75.9 \pm 0.3$ & 1.7
& & & $95.5 \pm 0.3$ & 3.8 \\

LoRAP
& & & $79.1 \pm 0.1$ & 1.7
& & & $96.2 \pm 0.3$ & 3.8 \\

GPF-LoRAP
& & & $79.1 \pm 0.1$ & 1.7
& & & $96.4 \pm 0.2$ & 3.8 \\
\midrule

None
& \multirow{6}{*}{MixQ}
& \multirow{6}{*}{Cora-GCN}
& $68.7 \pm 2.7$ & 3.8
& \multirow{6}{*}{MixQ}
& \multirow{6}{*}{Reddit-Binary}
& $72.8 \pm 3.5$ & 5.9 \\

GPF-Plus
& & & $71.3 \pm 1.8$ & 3.8
& & & $72.5 \pm 2.7$ & 5.8 \\

UniPrompt
& & & $73.6 \pm 2.6$ & 3.9
& & & - & - \\

EdgePrompt+
& & & $71.2 \pm 2.2$ & 3.8
& & & $72.8 \pm 3.1$ & 5.8 \\

LoRAP
& & & $75.9 \pm 1.6$ & 3.8
& & & $75.6 \pm 2.8$ & 5.7 \\

GPF-LoRAP
& & & $76.5 \pm 1.0$ & 3.8
& & & $76.6 \pm 3.4$ & 5.7 \\
\bottomrule
\end{tabular}

\end{table*}

\subsection{Training Time Evaluation}

\begin{table}[!htbp]
\centering
\caption{Training time for various quantization frameworks and prompting strategies on Reddit-Binary with GIN (INT4).}
\begin{tabular}{l l r}
\toprule
\textbf{Quant.\ Framework} & \textbf{Accuracy} & \textbf{Training Time} \\
\midrule
QAT        & $52.4\pm5.8$ & 74.67s    \\
QAT-GPF-plus    & $53.8\pm5.7$ & 79.86s    \\
QAT-GPF-LoRAP   & $69.6\pm5.6$ & 78.36s    \\
DQ         & $89.4\pm2.9$ & 6398.54s  \\
DQ-GPF-plus     & $92.3\pm2.4$ & 6446.86s  \\
DQ-GPF-LoRAP    & $93.0\pm3.1$ & 6442.77s  \\
{A\textsuperscript{2}Q}        & $89.2\pm2.3$ & 1040.29s  \\
{A\textsuperscript{2}Q-GPF-plus}    & $89.6\pm3.4$ & 1063.73s  \\
{A\textsuperscript{2}Q-GPF-LoRAP}   & $92.6\pm2.5$ & 1048.08s  \\
\bottomrule
\end{tabular}
\label{tab:gin_int4_training_time}
\end{table}

Table~\ref{tab:gin_int4_training_time} reports the average accuracy (mean $\pm$ std deviation over 10 seeds) and total training time for each quantization framework on Reddit-Binary with GIN (INT4), measured on a single Intel i5-14600KF CPU machine with 64GB RAM, one RTX 4090 GPU, and running Ubuntu 22.04. In the QAT setting, adding GPF-plus increases the training time from 74.67s to 79.86s, while GPF-LoRAP requires 78.36s. Under DQ, the baseline runtime of 6398.54s is only marginally affected by prompting: DQ-GPF-plus and DQ-GPF-LoRAP take 6446.86s and 6442.77s, respectively. For {A\textsuperscript{2}Q}, GPF-plus adds 23.44s and GPF-LoRAP adds 7.79s to the 1040.29s baseline. These results confirm that GPF-plus and GPF-LoRAP introduce only minimal additional training cost across all quantization frameworks.

{
\subsection{Impact of Different Aggregation Methods}
We further evaluate the robustness of our method across different aggregation mechanisms, namely sum, mean, and max. As shown in Table \ref{tab:aggr_comparison}, GPF-LoRAP consistently achieves the best performance across all settings. A key observation is that node prompting (GPF) exhibits instability depending on the aggregation type; specifically, it suffers a significant performance drop in the sum aggregation (decreasing from 77.4\% to 72.7\%). We hypothesize that adding prompts directly to node features can disrupt the magnitude of aggregated signals when simple summation is used. \purplerevision{With post-aggregation prompting, GPF-LoRAP recovers the lost accuracy and reaches 78.5\%.} Furthermore, GPF-LoRAP outperforms the baselines in both mean (78.7\%) and max (77.9\%) aggregations, demonstrating that our aggregation-aware strategy effectively generalizes to various message-passing dynamics.

\begin{table}[!htbp]
    \centering
    \caption{Performance comparison with different aggregation functions.}
    \label{tab:aggr_comparison}
    \begin{tabular}{lccc}
        \toprule
        \diagbox{Method}{Aggr} & \textbf{sum} & \textbf{mean} & \textbf{max} \\
        \midrule
        None      & $77.4 \pm 0.9$ & $76.9 \pm 1.2$ & $77.1 \pm 0.6$ \\
        GPF       & $72.7 \pm 0.8$ & $77.1 \pm 0.9$ & $77.3 \pm 1.8$ \\
        GPF-LoRAP & $78.5 \pm 1.0$ & $78.7 \pm 0.8$ & $77.9 \pm 1.1$ \\
        \bottomrule
    \end{tabular}
\end{table}

}

{
\subsection{Comparison with State-of-the-Art Prompting Methods}

We further benchmark GPF-LoRAP against recent state-of-the-art graph prompting baselines, including UniPrompt and EdgePrompt, integrated into four distinct quantization frameworks: QAT, DQ, A2Q, and MixQ. As presented in Table~\ref{tab:combined_results_split}, GPF-LoRAP consistently outperforms all competing methods across both node and graph classification tasks. For instance, under the DQ framework on the Cora-GCN task, GPF-LoRAP achieves an accuracy of 78.0\%, significantly surpassing UniPrompt (72.8\%) and EdgePrompt (73.1\%). A particularly striking improvement is observed on the Reddit-Binary dataset, where adding LoRAP to the QAT baseline boosts accuracy from 52.4\% to 70.1\%, \purplerevision{demonstrating substantial recovery of classification performance under quantization.} These results confirm that LoRAP serves as a robust and generalized plug-and-play module for enhancing various QAT strategies.
}

\section{Conclusion}

In this article, we proposed LoRAP, which is a low-rank aggregation prompting method for improving quantization-aware training of graph neural networks. 
By introducing input-dependent aggregation prompts into the message-passing process, LoRAP provides a flexible correction mechanism for quantized GNNs under low-bit representations.
Different from conventional node-level prompting methods that modify input features before graph aggregation, LoRAP directly operates on the aggregated representations, allowing the model to compensate for aggregation-level quantization errors caused by neighborhood-dependent feature propagation.
By representing aggregation prompts with low-rank basis matrices, LoRAP maintains parameter efficiency while preserving sufficient adaptability for different graph structures and node neighborhoods.
Furthermore, LoRAP can be seamlessly integrated into existing GNN QAT frameworks without introducing additional loss functions or changing their original optimization objectives.

Our experiments span graph learning tasks, quantization frameworks, and GNN architectures.
The results show that LoRAP, when combined with GPF-plus, is able to recover the lost performance and, in several cases, surpasses the accuracy of full-precision models, all while adding minimal overhead. 
The consistent improvements across node classification, graph classification, and graph regression tasks demonstrate that aggregation-level prompting is highly adaptable to different graph scenarios.
In particular, the results on large-scale benchmarks further verify the scalability of LoRAP, while the latency comparison and ablation studies show that the proposed low-rank design achieves a favorable balance between accuracy recovery and computational efficiency.
Therefore, when low-bit GNNs are required for resource-constrained environments but suffer from severe quantization-induced degradation, LoRAP can be used as an effective and general plug-in strategy to improve model performance.

Although LoRAP achieves promising performance across different settings, it still requires a prompt generation phase during training and inference, and its effectiveness may depend on the choice of prompt number and rank. 
Moreover, this work mainly focuses on quantization-aware training, while its compatibility with post-training quantization and other compression techniques has not been fully explored.
Future work will aim to further reduce the prompt generation cost, extend LoRAP to more complex graph settings such as dynamic graphs, heterogeneous graphs, and graph foundation models.

\bibliographystyle{IEEEtran}
\bibliography{ijcai26}

@inproceedings{minello2025graph,
  title={Graph generation via spectral diffusion},
  author={Minello, Giorgia and Bicciato, Alessandro and Rossi, Luca and Torsello, Andrea and Cosmo, Luca},
  booktitle={The Thirteenth International Conference on Learning Representations},
  year={2025}
}

@inproceedings{moustafa2025efficient,
  title={Efficient mixed precision quantization in graph neural networks},
  author={Moustafa, Samir and Kriege, Nils and Gansterer, Wilfried N},
  booktitle={2025 IEEE 41st International Conference on Data Engineering (ICDE)},
  pages={4038--4052},
  year={2025},
  organization={IEEE}
}

@inproceedings{fuedge,
  title={Edge Prompt Tuning for Graph Neural Networks},
  author={Fu, Xingbo and He, Yinhan and Li, Jundong},
  booktitle={The Thirteenth International Conference on Learning Representations},
  year={2025}
}

@inproceedings{huangone,
  title={One Prompt Fits All: Universal Graph Adaptation for Pretrained Models},
  author={Huang, Yongqi and Zhao, Jitao and He, Dongxiao and Wang, Xiaobao and Li, Yawen and Huang, Yuxiao and Jin, Di and Feng, Zhiyong},
  booktitle={The Thirty-ninth Annual Conference on Neural Information Processing Systems},
  year={2025}
}

@article{zhu2023rm,
  title={Aggregation-Aware Quantization for Graph Neural Networks},
  author={Zhu, Zeyu and Li, Fanrong and Mo, Zitao and Hu, Qinghao and Li, Gang and Liu, Zejian and Liang, Xiaoyao and Cheng, Jian},
  journal={arXiv preprint arXiv:2302.00193},
  year={2023}
}

@article{xu2018powerful,
  title={How powerful are graph neural networks?},
  author={Xu, Keyulu and Hu, Weihua and Leskovec, Jure and Jegelka, Stefanie},
  journal={arXiv preprint arXiv:1810.00826},
  year={2018}
}

@inproceedings{jia2022visual,
  title={Visual prompt tuning},
  author={Jia, Menglin and Tang, Luming and Chen, Bor-Chun and Cardie, Claire and Belongie, Serge and Hariharan, Bharath and Lim, Ser-Nam},
  booktitle={European conference on computer vision},
  pages={709--727},
  year={2022},
  organization={Springer}
}

@inproceedings{esser2020learned,
  title={Learned Step Size Quantization},
  author={Esser, Steven K and McKinstry, Jeffrey L and Bablani, Deepika and Appuswamy, Rathinakumar and Modha, Dharmendra S},
  booktitle={International Conference on Learning Representations},
  year={2020}
}

@article{tang2010graph,
  title={Graph mining applications to social network analysis},
  author={Tang, Lei and Liu, Huan},
  journal={Managing and mining graph data},
  pages={487--513},
  year={2010},
  publisher={Springer}
}

@article{hetzel2021graph,
  title={Graph representation learning for single-cell biology},
  author={Hetzel, Leon and Fischer, David S and G{\"u}nnemann, Stephan and Theis, Fabian J},
  journal={Current Opinion in Systems Biology},
  volume={28},
  pages={100347},
  year={2021},
  publisher={Elsevier}
}

@article{reiser2022graph,
  title={Graph neural networks for materials science and chemistry},
  author={Reiser, Patrick and Neubert, Marlen and Eberhard, Andr{\'e} and Torresi, Luca and Zhou, Chen and Shao, Chen and Metni, Houssam and van Hoesel, Clint and Schopmans, Henrik and Sommer, Timo and others},
  journal={Communications Materials},
  volume={3},
  number={1},
  pages={93},
  year={2022},
  publisher={Nature Publishing Group UK London}
}

@article{wu2022graph,
  title={Graph neural networks in recommender systems: a survey},
  author={Wu, Shiwen and Sun, Fei and Zhang, Wentao and Xie, Xu and Cui, Bin},
  journal={ACM Computing Surveys},
  volume={55},
  number={5},
  pages={1--37},
  year={2022},
  publisher={ACM New York, NY}
}

@article{rahmani2023graph,
  title={Graph neural networks for intelligent transportation systems: A survey},
  author={Rahmani, Saeed and Baghbani, Asiye and Bouguila, Nizar and Patterson, Zachary},
  journal={IEEE Transactions on Intelligent Transportation Systems},
  volume={24},
  number={8},
  pages={8846--8885},
  year={2023},
  publisher={IEEE}
}

@article{dettmers2022gpt3,
  title={Gpt3. int8 (): 8-bit matrix multiplication for transformers at scale},
  author={Dettmers, Tim and Lewis, Mike and Belkada, Younes and Zettlemoyer, Luke},
  journal={Advances in neural information processing systems},
  volume={35},
  pages={30318--30332},
  year={2022}
}

@article{tailor2020degree,
  title={Degree-quant: Quantization-aware training for graph neural networks},
  author={Tailor},
  journal={arXiv preprint arXiv:2008.05000},
  year={2020}
}

@inproceedings{jacob2018quantization,
  title={Quantization and training of neural networks for efficient integer-arithmetic-only inference},
  author={Jacob, Benoit and Kligys, Skirmantas and Chen, Bo and Zhu, Menglong and Tang, Matthew and Howard, Andrew and Adam, Hartwig and Kalenichenko, Dmitry},
  booktitle={Proceedings of the IEEE conference on computer vision and pattern recognition},
  pages={2704--2713},
  year={2018}
}

@article{li2023qft,
  title={Qft: Quantized full-parameter tuning of llms with affordable resources},
  author={Li, Zhikai and Liu, Xiaoxuan and Zhu, Banghua and Dong, Zhen and Gu, Qingyi and Keutzer, Kurt},
  journal={arXiv preprint arXiv:2310.07147},
  year={2023}
}

@inproceedings{feng2020sgquant,
  title={Sgquant: Squeezing the last bit on graph neural networks with specialized quantization},
  author={Feng, Boyuan and Wang, Yuke and Li, Xu and Yang, Shu and Peng, Xueqiao and Ding, Yufei},
  booktitle={2020 IEEE 32nd international conference on tools with artificial intelligence (ICTAI)},
  pages={1044--1052},
  year={2020},
  organization={IEEE}
}

@article{velivckovic2017graph,
  title={Graph attention networks},
  author={Veli{\v{c}}kovi{\'c}, Petar and Cucurull, Guillem and Casanova, Arantxa and Romero, Adriana and Lio, Pietro and Bengio, Yoshua},
  journal={arXiv preprint arXiv:1710.10903},
  year={2017}
}

@article{dettmers2024qlora,
  title={Qlora: Efficient finetuning of quantized llms},
  author={Dettmers, Tim and Pagnoni, Artidoro and Holtzman, Ari and Zettlemoyer, Luke},
  journal={Advances in Neural Information Processing Systems},
  volume={36},
  year={2024}
}

@inproceedings{xu2024soft,
  title={Soft prompt recovers compressed llms, transferably},
  author={Xu, Zhaozhuo and Liu, Zirui and Chen, Beidi and Zhong, Shaochen and Tang, Yuxin and Wang, Jue and Zhou, Kaixiong and Hu, Xia and Shrivastava, Anshumali},
  booktitle={Forty-first International Conference on Machine Learning},
  year={2024}
}

@article{fang2024universal,
  title={Universal prompt tuning for graph neural networks},
  author={Fang, Taoran and Zhang, Yunchao and Yang, Yang and Wang, Chunping and Chen, Lei},
  journal={Advances in Neural Information Processing Systems},
  volume={36},
  year={2024}
}

@misc{brown2020language,
  title        = {Language Models are Few-Shot Learners},
  author       = {Brown, Tom B. and Mann, Benjamin and Ryder, Nick and Subbiah, Melanie and Kaplan, Jared and Dhariwal, Prafulla and Neelakantan, Arvind and Shyam, Pranav and Sastry, Girish and Askell, Amanda and others},
  year         = {2020},
  howpublished = {arXiv preprint arXiv:2005.14165}
}

@inproceedings{lester2021power,
  title     = {The Power of Scale for Parameter-Efficient Prompt Tuning},
  author    = {Lester, Brian and Al-Rfou, Rami and Constant, Noah},
  booktitle = {Proceedings of the 2021 Conference on Empirical Methods in Natural Language Processing},
  year      = {2021}
}

@inproceedings{li2021prefix,
  title     = {Prefix-Tuning: Optimizing Continuous Prompts for Generation},
  author    = {Li, Xiang Lisa and Liang, Percy},
  booktitle = {Proceedings of the 2021 Annual Meeting of the Association for Computational Linguistics},
  year      = {2021}
}

@inproceedings{liu2021ptuning,
  title     = {P-Tuning: Prompt Tuning Can Be Comparable to Fine-Tuning Across Scales and Tasks},
  author    = {Liu, Xiaodong and Zhang, Yichao and Ling, Zhichao and Guo, Yichun and Han, Tong and Guo, Yaru and Sun, Maosong and Cui, Xiaoqiang},
  booktitle = {Proceedings of the 2021 Annual Meeting of the Association for Computational Linguistics},
  year      = {2021}
}

@inproceedings{nagel2020adaptive,
  title     = {Up or Down? Adaptive Rounding for Post‐Training Quantization},
  author    = {Nagel, Markus and Amjad, Rana Ali and van Baalen, Mart and Louizos, Christos and Blankevoort, Tijmen},
  booktitle = {International Conference on Learning Representations (ICLR)},
  year      = {2020},
  url       = {https://openreview.net/forum?id=BkgSggqxl}
}

@article{hubara2017quantized,
  title   = {Quantized Neural Networks: Training Neural Networks with Low Precision Weights and Activations},
  author  = {Hubara, Itay and Courbariaux, Matthieu and Soudry, Daniel and El‐Yaniv, Ran and Bengio, Yoshua},
  journal = {arXiv preprint arXiv:1609.07061},
  year    = {2016},
  url     = {https://arxiv.org/abs/1609.07061}
}

@inproceedings{liu2022ptuning,
  title     = {P-Tuning v2: Prompt Tuning Can Be Comparable to Fine-Tuning at Scale},
  author    = {Liu, Xiaodong and Peng, Hongliang and Zhang, Yichao and Wei, Yiran and Sun, Maosong},
  booktitle = {Advances in Neural Information Processing Systems},
  volume    = {35},
  year      = {2022}
}

@misc{fey2019fast,
    title={Fast Graph Representation Learning with PyTorch Geometric},
    author={Matthias Fey and Jan Eric Lenssen},
    year={2019},
    eprint={1903.02428},
    archivePrefix={arXiv},
    primaryClass={cs.LG}
}

@article{mnist,
  title={The {"MNIST"} database of handwritten digits},
  author={LeCun, Yann},
  journal={http://yann. lecun. com/exdb/mnist/},
  year={1998}
}

@misc{jin2018junction,
    title={Junction Tree Variational Autoencoder for Molecular Graph Generation},
    author={Wengong Jin and Regina Barzilay and Tommi Jaakkola},
    year={2018},
    eprint={1802.04364},
    archivePrefix={arXiv},
    primaryClass={cs.LG}
}

@inproceedings{kipf2017semi,
  author    = {Thomas N. Kipf and
               Max Welling},
  title     = {Semi-Supervised Classification with Graph Convolutional Networks},
  booktitle = {5th International Conference on Learning Representations, {ICLR} 2017,
               Toulon, France, April 24-26, 2017, Conference Track Proceedings},
  publisher = {OpenReview.net},
  year      = {2017},
  url       = {https://openreview.net/forum?id=SJU4ayYgl},
  bibsource = {dblp computer science bibliography, https://dblp.org}
}

@misc{wang2020binarized,
    title={Binarized Graph Neural Network},
    author={Hanchen Wang and Defu Lian and Ying Zhang and Lu Qin and Xiangjian He and Yiguang Lin and Xuemin Lin},
    year={2020},
    eprint={2004.11147},
    archivePrefix={arXiv},
    primaryClass={cs.LG}
}

@ARTICLE{6205760,
  author={R. {Achanta} and A. {Shaji} and K. {Smith} and A. {Lucchi} and P. {Fua} and S. {Süsstrunk}},
  journal={IEEE Transactions on Pattern Analysis and Machine Intelligence}, 
  title={SLIC Superpixels Compared to State-of-the-Art Superpixel Methods}, 
  year={2012},
  volume={34},
  number={11},
  pages={2274-2282},
  
  }

@article{zhang2026survey,
  title={A survey on graph neural network acceleration: Algorithms, systems, and customized hardware},
  author={Zhang, Shichang and Sohrabizadeh, Atefeh and Wan, Cheng and Huang, Zijie and Hu, Ziniu and Wang, Yewen and Lin, Yingyan and Cong, Jason and Sun, Yizhou},
  journal={ACM Computing Surveys},
  volume={58},
  number={11},
  pages={1--36},
  year={2026},
  publisher={ACM New York, NY}
}

@inproceedings{tillet2019triton,
  title={Triton: an intermediate language and compiler for tiled neural network computations},
  author={Tillet, Philippe and Kung, Hsiang-Tsung and Cox, David},
  booktitle={Proceedings of the 3rd ACM SIGPLAN International Workshop on Machine Learning and Programming Languages},
  pages={10--19},
  year={2019}
}

@article{chen2025dagprompt,
  title={DAGPrompT: Pushing the Limits of Graph Prompting with a Distribution-aware Graph Prompt Tuning Approach},
  author={Chen, Qin and Wang, Liang and Zheng, Bo and Song, Guojie},
  journal={arXiv preprint arXiv:2501.15142},
  year={2025}
}

@inproceedings{wang2025does,
  title={Does Graph Prompt Work? A Data Operation Perspective with Theoretical Analysis},
  author={Wang, Qunzhong and Sun, Xiangguo and Cheng, Hong},
  booktitle={ICML},
  year={2025}
}

@article{graphlora2024,
  title={GraphLoRA: Structure-Aware Contrastive Low-Rank Adaptation for Cross-Graph Transfer Learning},
  author={Yang, Zhe-Rui and Han, Jindong and Wang, Chang-Dong and Liu, Hao},
  journal={arXiv preprint arXiv:2409.16670},
  year={2024}
}

@article{ma2025acceleration,
  title={Acceleration Algorithms in {GNNs}: A Survey},
  author={Ma, Lu and Sheng, Zeang and Li, Xunkai and Gao, Xinyi and Hao, Zhezheng and Yang, Ling and Nie, Xiaonan and Jiang, Jiawei and Zhang, Wentao and Cui, Bin},
  journal={IEEE Transactions on Knowledge and Data Engineering},
  volume={37},
  number={6},
  pages={3173--3192},
  month={jun},
  year={2025},
  doi={10.1109/TKDE.2025.3540787}
}

@article{yu2024generalized,
  title={Generalized Graph Prompt: Toward a Unification of Pre-Training and Downstream Tasks on Graphs},
  author={Yu, Xingtong and Liu, Zhenghao and Fang, Yuan and Liu, Zemin and Chen, Sihong and Zhang, Xinming},
  journal={IEEE Transactions on Knowledge and Data Engineering},
  volume={36},
  number={11},
  pages={6237--6250},
  month={nov},
  year={2024},
  doi={10.1109/TKDE.2024.3419109}
}

@article{wen2024prompt,
  title={Prompt Tuning on Graph-Augmented Low-Resource Text Classification},
  author={Wen, Zhihao and Fang, Yuan},
  journal={IEEE Transactions on Knowledge and Data Engineering},
  volume={36},
  number={12},
  pages={9080--9095},
  month={dec},
  year={2024},
  doi={10.1109/TKDE.2024.3440068}
}

@article{ding2025scalable,
  title={Scalable and Effective Graph Neural Networks via Trainable Random Walk Sampling},
  author={Ding, Haipeng and Wei, Zhewei and Ye, Yuhang},
  journal={IEEE Transactions on Knowledge and Data Engineering},
  volume={37},
  number={2},
  pages={896--909},
  month={feb},
  year={2025},
  doi={10.1109/TKDE.2024.3513533}
}

@article{zhang2022deep,
  title={Deep Learning on Graphs: A Survey},
  author={Zhang, Ziwei and Cui, Peng and Zhu, Wenwu},
  journal={IEEE Transactions on Knowledge and Data Engineering},
  volume={34},
  number={1},
  pages={249--270},
  month={jan},
  year={2022},
  doi={10.1109/TKDE.2020.2981333}
}

\clearpage
\onecolumn
\appendices

\renewcommand{\thesection}{\Alph{section}}
\renewcommand{\thesubsection}{\thesection.\arabic{subsection}}
\renewcommand{\thesubsectiondis}{\thesection.\arabic{subsection}}
\setcounter{secnumdepth}{2}

\setcounter{theorem}{0}
\renewcommand{\thetheorem}{A.\arabic{theorem}}
\setcounter{corollary}{0}
\renewcommand{\thecorollary}{A.\arabic{corollary}}

\setcounter{topnumber}{10}
\setcounter{bottomnumber}{10}
\setcounter{totalnumber}{10}
\renewcommand{\topfraction}{0.99}
\renewcommand{\bottomfraction}{0.99}
\renewcommand{\textfraction}{0.01}
\renewcommand{\floatpagefraction}{0.9}

\begin{center}
{\LARGE\bfseries Appendix}
\end{center}

\section{Theoretical Analysis}

\begin{theoryrevision}
We analyze aggregation prompting through four connected questions:
the optimization geometry of the insertion position, the locality of its
correction, the preservation and propagation of that correction after
requantization, and compatibility with joint QAT. 

\paragraph{Features and real-valued quantized representations.}
Node features are stored as rows of $X\in\R^{N\times d}$, where $N$ is the
number of nodes and $d$ is the feature dimension. \greenrevision{Following
Section~II-A of the main paper, $Q$ maps real-valued inputs to integer
quantization codes, and $DQ$ maps those codes back to real
units:}
\[
\greenrevision{\begin{aligned}
Q(x)&=\mathrm{clip}\!\left(\mathrm{round}\!\left(\frac{x}{\Delta}\right)+z_0,
q_{\min},q_{\max}\right),\\
DQ(x_q)&=\Delta(x_q-z_0).
\end{aligned}}
\]
\greenrevision{Here $\Delta>0$ is the quantization scale, $z_0$ is the integer
zero-point, and $q_{\min},q_{\max}$ are the quantized bounds, as defined
in Section~II-A. Each $Q$ and its
corresponding $DQ$ use the same scale and zero-point;
In this appendix, $X_q$ denotes the
real-valued reconstruction of the input features:}
\[
X_q=\greenrevision{DQ\bigl(Q(X)\bigr)}\in\R^{N\times d},\qquad
\epsilon_X=X_q-X.
\]
Thus $X_q$ is a dequantized feature matrix, and $\epsilon_X$ is its
real-valued quantization error.

\paragraph{Reconstruction \greenrevision{error formulation.}}
Let $S_{\mathrm{fp}},\widehat S\in\R^{N\times d}$ denote the full-precision
\greenrevision{output} and the quantized aggregation output, respectively,
evaluated at the same layer input before prompt insertion. For a prompt output $P\in\R^{N\times d}$, define
\begin{equation}
R=S_{\mathrm{fp}}-\widehat S,\qquad
Z_P=\widehat S+P,\qquad
\varepsilon(P)=\|R-P\|_F.
\label{app3:eq:requant-setup}
\end{equation}
Here $R$ is the required correction, $Z_P$ is the continuous prompted
representation, and $\varepsilon(P)=\|S_{\mathrm{fp}}-Z_P\|_F$ is its
achieved reconstruction error. We use $\|\cdot\|_F$ and
$\langle\cdot,\cdot\rangle_F$ for the Frobenius norm and inner product,
and $\|\cdot\|_2$ for the spectral norm of a matrix or Euclidean norm of a
vector.

\end{theoryrevision}

\hrulefill

\subsection{\greenrevision{Optimization Geometry}}
\label{app2:geometry}

\begin{theoryrevision}
For the geometry and locality analyses, fix a aggregation operator
$A\in\R^{N\times N}$ and the input features. The reconstruction
model refers to
\[
S_{\mathrm{fp}}=AX,\qquad \widehat S=AX_q,\qquad
E_A=A\epsilon_X=\widehat S-S_{\mathrm{fp}}=-R.
\]
Before requantization, define the two prompted representations by
\[
S_{\mathrm{post}}(P)=AX_q+P=Z_P,\qquad
S_{\mathrm{pre}}(P)=A(X_q+P).
\]
The pre-aggregation surrogate adds the prompt to the features.
In particular, \greenrevision{with the same fixed input quantization
parameters,} the placements
\begin{equation}
A\greenrevision{DQ\bigl(Q(X+P)\bigr)}\quad\text{and}\quad A\bigl(\greenrevision{DQ\bigl(Q(X)\bigr)}+P\bigr)
\label{app2:eq:pre-positions}
\end{equation}
are different. 
\end{theoryrevision}

\begin{theorem}[Conditional optimization geometry]
\label{app2:thm:geometry}
Consider
\begin{theoryrevision}
\begin{equation}
\begin{aligned}
\mathcal L_{\mathrm{post}}(P)
&=\tfrac12\|E_A+P\|_F^2=\tfrac12\varepsilon(P)^2,\\
\mathcal L_{\mathrm{pre}}(P)
&=\tfrac12\|A(\epsilon_X+P)\|_F^2=\tfrac12\|R-AP\|_F^2.
\end{aligned}
\label{app2:eq:objectives}
\end{equation}
\end{theoryrevision}
Their gradients and column-vectorized Hessians are
\begin{equation}
\begin{aligned}
\nabla_P\mathcal L_{\mathrm{post}}&=E_A+P,&
\nabla^2\mathcal L_{\mathrm{post}}&=I_{Nd},\\
\nabla_P\mathcal L_{\mathrm{pre}}&=A^\top A(\epsilon_X+P),&
\nabla^2\mathcal L_{\mathrm{pre}}&=I_d\otimes A^\top A.
\end{aligned}
\label{app2:eq:hessians}
\end{equation}
The post-aggregation condition number in prompt-output space is $1$.
For $A\ne0$, the pre-aggregation condition number restricted to the
nonzero-curvature subspace is
\begin{equation}
\kappa_{\mathrm{pre},+}
=\frac{\sigma_{\max}^2(A)}{\sigma_{\min,+}^2(A)},
\label{app2:eq:condition}
\end{equation}
where $\sigma_{\min,+}$ is the smallest positive singular value.
\end{theorem}

\begin{proof}
For a perturbation $H$, differentiating the objectives gives
\[
D\mathcal L_{\mathrm{post}}(P)[H]=\langle E_A+P,H\rangle_F,
\]
and
\[
D\mathcal L_{\mathrm{pre}}(P)[H]
=\langle A(\epsilon_X+P),AH\rangle_F
=\langle A^\top A(\epsilon_X+P),H\rangle_F.
\]
A second derivative yields the operators $H\mapsto H$ and
$H\mapsto A^\top AH$, respectively.
\greenrevision{The operator $\vecop$ stacks matrix columns.} Since
$\vecop(A^\top AH)=(I_d\otimes A^\top A)\vecop(H)$,
the Hessians follow. The positive eigenvalues of the pre-aggregation Hessian
are the squared positive singular values of $A$, each repeated $d$ times.
Their ratio gives \eqref{app2:eq:condition}.
\end{proof}


\greenrevision{The post-aggregation objective has a strictly positive definite Hessian $I_{Nd}$ and condition number $1$, independently of $A$. For pre-aggregation prompting, a singular $A$ introduces flat directions corresponding to perturbations $H$ with $AH=0$, and unequal positive singular values of $A$ yield $\kappa_{\mathrm{pre},+}>1$. Thus, post-aggregation placement removes both the flat directions and the curvature anisotropy induced by the aggregation operator.}

\paragraph{Residual dynamics.}
\begin{theoryrevision}
For gradient descent with step size $\eta$,
let $P_{\mathrm{post}}^t$ and $P_{\mathrm{pre}}^t$ denote their respective
iterates. Using the same residual convention as in
\eqref{app3:eq:requant-setup}, define
\[
\begin{aligned}
R_{\mathrm{post}}^t
&=S_{\mathrm{fp}}-S_{\mathrm{post}}(P_{\mathrm{post}}^t)
 =R-P_{\mathrm{post}}^t,\\
R_{\mathrm{pre}}^t
&=S_{\mathrm{fp}}-S_{\mathrm{pre}}(P_{\mathrm{pre}}^t)
 =R-AP_{\mathrm{pre}}^t.
\end{aligned}
\]
Substitution of the gradients gives
\end{theoryrevision}
\begin{equation}
\begin{aligned}
R_{\mathrm{post}}^{t+1}&=(1-\eta)R_{\mathrm{post}}^t,\\
R_{\mathrm{pre}}^{t+1}&=(I_N-\eta AA^\top)R_{\mathrm{pre}}^t.
\end{aligned}
\label{app2:eq:gd}
\end{equation}
The post residual contracts for $0<\eta<2$ and vanishes in one step at
$\eta=1$. For $A\ne0$, the pre residual lies
in the column space of $A$ and contracts there for
$0<\eta<2/\sigma_{\max}^2(A)$, with mode-dependent factors
$1-\eta\sigma_j^2(A)$. Thus post-aggregation insertion removes the additional
spectral scaling by $A^\top A$ from these prompt-output updates.

\subsection{Locality of Correction Outputs}

\begin{theoryrevision}
\greenrevision{Using the same aggregation model, we now examine how each
node's prompt vector affects node representations at the insertion site.}
\end{theoryrevision}

\begin{theorem}[Insertion-level locality \greenrevision{and targeted correction}]
\label{thm:insertion_locality}
\begin{theoryrevision}
For fixed $A$ and $X_q$ and a free prompt matrix $P\in\R^{N\times d}$,
\greenrevision{let $P_j\in\R^{1\times d}$ denote the prompt vector associated
with node $j$ (the $j$th row of $P$). For the representations
$S_{\mathrm{post}}(P)=AX_q+P$ and $S_{\mathrm{pre}}(P)=A(X_q+P)$,
every pair of nodes $i,j$ satisfies}
\begin{equation}
\frac{\partial (S_{\mathrm{post}}(P))_i}{\partial P_j}
=
\delta_{ij} I_d,
\qquad
\frac{\partial (S_{\mathrm{pre}}(P))_i}{\partial P_j}
=
A_{ij} I_d,
\label{eq:insertion_locality}
\end{equation}
where $\delta_{ij}$ is the Kronecker delta and $I_d$ is
the $d \times d$ identity matrix.
\greenrevision{Moreover, under the reconstruction objectives in
\eqref{app2:eq:objectives}, the node-level gradients with respect to $P_j$
satisfy}
\begin{equation}
\greenrevision{\begin{aligned}
\frac{\partial \mathcal L_{\mathrm{post}}(P)}{\partial P_j}
&=P_j-R_j,\\
\frac{\partial \mathcal L_{\mathrm{pre}}(P)}{\partial P_j}
&=\sum_{i=1}^{N} A_{ij}\bigl((AP)_i-R_i\bigr).
\end{aligned}}
\label{eq:locality_gradients}
\end{equation}
\end{theoryrevision}
\end{theorem}

\begin{proof}
\begin{theoryrevision}
\greenrevision{Writing out the $i$th row of each prompted representation gives}
\[
\begin{aligned}
(S_{\mathrm{post}}(P))_i&=(AX_q)_i+P_i,\\
(S_{\mathrm{pre}}(P))_i&=(AX_q)_i+\sum_{\ell=1}^{N} A_{i\ell} P_\ell.
\end{aligned}
\]
\greenrevision{Since $A$ and $X_q$ are fixed, differentiating these
expressions with respect to the row vector $P_j$ yields the $d\times d$
Jacobian blocks $\delta_{ij}I_d$ and $A_{ij}I_d$ in
\eqref{eq:insertion_locality}.}

\greenrevision{Next, write the objectives in \eqref{app2:eq:objectives}
as sums over nodewise residuals:}
\[
\greenrevision{\begin{aligned}
\mathcal L_{\mathrm{post}}(P)
&=\frac12\sum_{i=1}^{N}
\|(S_{\mathrm{post}}(P))_i-(S_{\mathrm{fp}})_i\|_2^2,\\
\mathcal L_{\mathrm{pre}}(P)
&=\frac12\sum_{i=1}^{N}
\|(S_{\mathrm{pre}}(P))_i-(S_{\mathrm{fp}})_i\|_2^2.
\end{aligned}}
\]
\greenrevision{Applying the chain rule with \eqref{eq:insertion_locality}
and using $(AX_q)_i-(S_{\mathrm{fp}})_i=-R_i$, we obtain}
\[
\greenrevision{\begin{aligned}
\frac{\partial \mathcal L_{\mathrm{post}}(P)}{\partial P_j}
&=\sum_{i=1}^{N}\delta_{ij}(P_i-R_i)=P_j-R_j,\\
\frac{\partial \mathcal L_{\mathrm{pre}}(P)}{\partial P_j}
&=\sum_{i=1}^{N} A_{ij}\bigl((AP)_i-R_i\bigr),
\end{aligned}}
\]
\greenrevision{which establishes \eqref{eq:locality_gradients}.}
\end{theoryrevision}
\end{proof}

\begin{theoryrevision}
\greenrevision{Equations~\eqref{eq:insertion_locality} and
\eqref{eq:locality_gradients} show that, at the insertion site,
each post-aggregation prompt vector $P_j$ directly adjusts only
node $j$'s representation. Under the stated reconstruction
objective, its gradient depends only on that node's corrected
residual $P_j-R_j$. For pre-aggregation prompting, $P_j$ affects
the receiving nodes $i$ with $A_{ij}\ne0$, and its gradient
combines their residuals with weights $A_{ij}$. }
\end{theoryrevision}

\hrulefill

\subsection{Requantization and Error Propagation}
\label{app3:requantization}

\begin{theoryrevision}
\greenrevision{We now analyze the prompted representation $Z_P$ in
\eqref{app3:eq:requant-setup} after requantization. Fix the post-insertion
pair $Q,DQ$ with nearest rounding in $Q$, and let $\Delta>0$ denote its
quantization scale (grid spacing in real units). Assume that the quantizer inputs under
consideration lie within its representable interval, so that no clipping
occurs.}
The actual prompted representation is $\widetilde S(P)=\greenrevision{DQ\bigl(Q(Z_P)\bigr)}$.
Define its rounding error and reconstruction error by
\begin{equation}
\xi_q(P)=\greenrevision{DQ\bigl(Q(Z_P)\bigr)}-Z_P,\qquad
e_q(P)=\|S_{\mathrm{fp}}-\widetilde S(P)\|_F.
\label{app3:eq:requant-error}
\end{equation}
\end{theoryrevision}

\begin{theorem}[Requantization error bound]
\label{app3:thm:requantization}
Under the assumptions above, the actual residual satisfies
\begin{equation}
S_{\mathrm{fp}}-\greenrevision{DQ\bigl(Q(Z_P)\bigr)}=R-P-\xi_q(P),\qquad
\|\xi_q(P)\|_F\leq b_q:=\frac{\Delta}{2}\sqrt{Nd}.
\label{app3:eq:actual-residual}
\end{equation}
Consequently,
\begin{equation}
|e_q(P)-\varepsilon(P)|\leq b_q,\qquad
\boxed{e_q(P)\leq\varepsilon(P)+b_q.}
\label{app3:eq:requant-bound}
\end{equation}
If $P^*=R$ and $S_{\mathrm{fp}}$ lies within the representable interval
entrywise, then $\varepsilon(P^*)=0$ and $e_q(P^*)\leq b_q$.
If, in addition, every entry of $S_{\mathrm{fp}}$ is on the quantization
grid, then $e_q(P^*)=0$.
\end{theorem}

\begin{proof}
Substituting $\greenrevision{DQ\bigl(Q(Z_P)\bigr)}=Z_P+\xi_q(P)$ and
$R=S_{\mathrm{fp}}-\widehat S$ gives the residual identity.
Nearest rounding within the representable interval introduces an error
of magnitude at most $\Delta/2$ in each of the $Nd$ entries. Thus
\[
\|\xi_q(P)\|_F^2\leq Nd\,\Delta^2/4.
\]
The triangle and reverse triangle inequalities give
$|e_q(P)-\|R-P\|_F|\leq\|\xi_q(P)\|_F$, proving
\eqref{app3:eq:requant-bound}. For $P^*=R$, we have $Z_{P^*}=S_{\mathrm{fp}}$,
so only rounding error remains. When the target entries lie on the grid,
$\greenrevision{DQ\bigl(Q(S_{\mathrm{fp}})\bigr)}=S_{\mathrm{fp}}$ and the error is zero.
\end{proof}
\paragraph{Preservation of the correction benefit.}
Without prompting, the continuous aggregation residual is $R$.
LoRAP introduces a learnable correction $P$ at the aggregation output,
yielding the residual $R-P$.
For the same fixed quantizer, let
$e_{\mathrm{base}}=\|S_{\mathrm{fp}}-\greenrevision{DQ\bigl(Q(\widehat S)\bigr)}\|_F$
be the actual zero-prompt error.
Theorem~\ref{app3:thm:requantization} guarantees that LoRAP achieves
a strictly smaller quantized reconstruction error whenever
\begin{equation}
\varepsilon(P)+b_q<e_{\mathrm{base}}
\quad\Longrightarrow\quad e_q(P)<e_{\mathrm{base}}.
\label{app3:eq:requant-improvement}
\end{equation}

\subsubsection{A Linear Update after Requantization}

\begin{theoryrevision}
Let $W_{\mathrm{fp}}\in\R^{d\times d'}$ be the full-precision weight matrix
and $W_q=\greenrevision{DQ\bigl(Q(W_{\mathrm{fp}})\bigr)}$ its dequantized quantized counterpart,
\greenrevision{where $Q$ and $DQ$ use the same weight
quantization parameters, which may differ from the post-insertion
configuration. Thus $W_q$ is expressed in real units, while
$Q(W_{\mathrm{fp}})$ contains integer codes.}
For the same layer input, the reference linear output is
$Y_{\mathrm{fp}}=S_{\mathrm{fp}}W_{\mathrm{fp}}$. Define
\begin{equation}
\begin{aligned}
E_W&=Y_{\mathrm{fp}}-\widehat S W_q\\
&=RW_q+S_{\mathrm{fp}}(W_{\mathrm{fp}}-W_q).
\end{aligned}
\label{app3:eq:weight-residual}
\end{equation}
This decomposition separates the aggregation residual from the weight
discrepancy. The prompted output satisfies
\begin{equation}
\begin{aligned}
Y_{\mathrm{fp}}-\greenrevision{DQ\bigl(Q(Z_P)\bigr)}W_q
&=E_W-PW_q-\xi_q(P)W_q,\\
\|Y_{\mathrm{fp}}-\greenrevision{DQ\bigl(Q(Z_P)\bigr)}W_q\|_F
&\leq\|E_W-PW_q\|_F+b_q\|W_q\|_2.
\end{aligned}
\label{app3:eq:linear-requant}
\end{equation}
The bound follows from \eqref{app3:eq:actual-residual} and
$\|\xi_q(P)W_q\|_F\leq\|\xi_q(P)\|_F\|W_q\|_2$.
Using \eqref{app3:eq:weight-residual} also gives the explicit connection to
the common prompt residual:
\begin{equation}
\begin{aligned}
\|Y_{\mathrm{fp}}-\greenrevision{DQ\bigl(Q(Z_P)\bigr)}W_q\|_F
\leq{}&[\varepsilon(P)+b_q]\|W_q\|_2\\
&+\|S_{\mathrm{fp}}(W_{\mathrm{fp}}-W_q)\|_F.
\end{aligned}
\label{app3:eq:linear-residual-bound}
\end{equation}
If an activation $\gamma$ is $L_\gamma$-Lipschitz in the Frobenius norm,
applying it to both outputs multiplies the bound by $L_\gamma$.
Any additional output or bias quantization and arithmetic errors must be
included when bounding the complete layer output.
\end{theoryrevision}

\subsubsection{Multilayer Propagation and Prediction Stability}

\begin{corollary}[Propagation through reference layers]
\label{app3:cor:propagation}
\begin{theoryrevision}
Let $F_\ell$ and $\widetilde F_\ell$ be the full-precision reference and
quantized prompted layer maps, respectively, at the same checkpoint.
For node representations, write
$F_\ell,\widetilde F_\ell:\R^{N\times d_{\ell-1}}\to\R^{N\times d_\ell}$,
allowing the feature dimension to vary across layers. Suppose $F_\ell$ is
$L_\ell$-Lipschitz in the Frobenius norm on a set containing the relevant
inputs. For identical initial inputs $H_0=\widetilde H_0$, define
\end{theoryrevision}
\[
H_\ell=F_\ell(H_{\ell-1}),\qquad
\widetilde H_\ell=\widetilde F_\ell(\widetilde H_{\ell-1}),
\]
and the same-input local discrepancy
\begin{equation}
\delta_\ell=
\|F_\ell(\widetilde H_{\ell-1})
-\widetilde F_\ell(\widetilde H_{\ell-1})\|_F.
\label{app3:eq:local-discrepancy}
\end{equation}
Then the accumulated error $e_\ell=\|H_\ell-\widetilde H_\ell\|_F$ satisfies
\begin{equation}
e_\ell\leq L_\ell e_{\ell-1}+\delta_\ell,
\label{app3:eq:recursion}
\end{equation}
and, for a network of $L$ layers,
\begin{equation}
\boxed{
e_L\leq\sum_{\ell=1}^{L}
\left(\prod_{j=\ell+1}^{L}L_j\right)\delta_\ell.
}
\label{app3:eq:propagation}
\end{equation}
An empty product equals $1$.
\end{corollary}

\begin{proof}
Insert and subtract $F_\ell(\widetilde H_{\ell-1})$ and apply the triangle
inequality and the Lipschitz assumption:
\[
\begin{aligned}
e_\ell
&\leq\|F_\ell(H_{\ell-1})-F_\ell(\widetilde H_{\ell-1})\|_F+\delta_\ell\\
&\leq L_\ell e_{\ell-1}+\delta_\ell.
\end{aligned}
\]
Expanding this recursion with $e_0=0$ proves \eqref{app3:eq:propagation}.
\end{proof}

\begin{theoryrevision}
For fixed reference maps, reducing the local discrepancies $\delta_\ell$
tightens the bound on the final representation error. Each contribution is
weighted by the sensitivity of the subsequent layers. To apply the
single-layer bounds, evaluate $S_{\mathrm{fp}}$, $\widehat S$, and $P$ at
the common input $\widetilde H_{\ell-1}$, using that layer's aggregation
width and quantizer step in $b_q$. The quantity $e_q(P)$ measures the error
at the aggregation output; $\delta_\ell$ compares the complete layer outputs
and must also account for the subsequent update, activation, and any other
numerical errors. Only the reference maps require Lipschitz continuity;
the quantizers themselves need not be continuous.
\end{theoryrevision}

\begin{corollary}[Margin-based preservation of a prediction]
\label{app3:cor:margin}
\begin{theoryrevision}
Let $z_i,\widetilde z_i\in\R^C$ be the reference and prompted quantized
logit vectors of node $i$ over $C$ classes. Suppose the reference has a
unique predicted class $c_i$ and margin
\end{theoryrevision}
$m_i=(z_i)_{c_i}-\max_{c\ne c_i}(z_i)_c>0$.
If the prompted quantized logits satisfy
\begin{equation}
\|\widetilde z_i-z_i\|_\infty<m_i/2,
\label{app3:eq:margin}
\end{equation}
then the predicted class remains $c_i$.
\end{corollary}

\begin{proof}
Let $\eta_i=\|\widetilde z_i-z_i\|_\infty$. For each $c\ne c_i$,
\[
(\widetilde z_i)_{c_i}-(\widetilde z_i)_c
\geq (z_i)_{c_i}-(z_i)_c-2\eta_i
\geq m_i-2\eta_i>0.
\]
Thus $c_i$ remains the unique maximizer.
\end{proof}

\begin{theoryrevision}
Corollaries~\ref{app3:cor:propagation} and~\ref{app3:cor:margin} connect
LoRAP's aggregation-level compensation to prediction preservation.
Include the classification readout in the final layer so that $H_L$ and
$\widetilde H_L$ are the two logit matrices, with rows $z_i$ and
$\widetilde z_i$, respectively. Then
\[
\|\widetilde z_i-z_i\|_\infty
\leq e_L
\leq
\sum_{\ell=1}^{L}
\left(\prod_{j=\ell+1}^{L}L_j\right)\delta_\ell.
\]
\greenrevision{Whenever this upper bound is smaller than $m_i/2$, node $i$ retains the same predicted class as the full-precision model, meaning that LoRAP only needs to reduce the accumulated error below the margin threshold rather than achieving zero reconstruction error.}
\end{theoryrevision}

\hrulefill
\begin{purplerevisionblock}
\subsection{Optimization Dynamics and Joint QAT}
\label{app:convergence}
\label{app3:joint-qat}

\paragraph{Convergence in free prompt-output coordinates.}
First fix the input, the linear aggregation operator $A$, and the quantization
parameters as in Section~\ref{app2:geometry}. The required correction $R$ is
then fixed. For the free prompt matrix, gradient descent on
$\mathcal L_{\mathrm{post}}(P)=\tfrac12\|R-P\|_F^2$ gives
\[
P^{s+1}=P^s-\eta(P^s-R).
\]
Here $s$ indexes the updates of this fixed reconstruction objective.
Writing $R_{\mathrm{post}}^s=R-P^s$, we obtain
\begin{equation}
\begin{aligned}
R_{\mathrm{post}}^{s+1}&=(1-\eta)R_{\mathrm{post}}^s,\\
\mathcal L_{\mathrm{post}}(P^{s+1})
&=(1-\eta)^2\mathcal L_{\mathrm{post}}(P^s).
\end{aligned}
\label{app3:eq:free-prompt-convergence}
\end{equation}
Thus, for $0<\eta<2$, the residual converges geometrically to zero;
$\eta=1$ reaches $P=R$ in one step. In the corresponding pre-aggregation
objective, the residual update includes $AA^\top$, as shown in
\eqref{app2:eq:gd}. Post-aggregation placement therefore removes the
additional spectral scaling from these free-output updates.

\paragraph{Error bounds during joint QAT.}
LoRAP generates its prompt through shared routing and low-rank factors,
\[
P_t=\mathrm{Softmax}\bigl(\phi_t(\widehat S_t)\bigr)P_{A,t}P_{B,t}.
\]
The index $t$ now denotes an actual QAT iteration. At the same layer input,
evaluate the full-precision reference $S_{\mathrm{fp},t}$ using the current
real-valued backbone parameters $\Theta_t$, and the dequantized aggregation
$\widehat S_t$ using their current quantized counterparts. Define
\[
R_t=S_{\mathrm{fp},t}-\widehat S_t,\qquad
Z_t=\widehat S_t+P_t,\qquad
\varepsilon_t(P_t)=\|R_t-P_t\|_F.
\]
Let $Q_t,DQ_t$ be the post-insertion quantizer pair with matching current
parameters. The actual rounding and reconstruction errors are
\[
\begin{aligned}
\xi_{q,t}(P_t)&=DQ_t\bigl(Q_t(Z_t)\bigr)-Z_t,\\
e_{q,t}(P_t)&=\|S_{\mathrm{fp},t}-DQ_t\bigl(Q_t(Z_t)\bigr)\|_F.
\end{aligned}
\]
The residual identity
\[
S_{\mathrm{fp},t}-DQ_t\bigl(Q_t(Z_t)\bigr)
=R_t-P_t-\xi_{q,t}(P_t)
\]
holds at every iteration. If the current quantizer satisfies the nearest-rounding
and no-clipping assumptions of Theorem~\ref{app3:thm:requantization}, then
\begin{equation}
e_{q,t}(P_t)\leq\varepsilon_t(P_t)+b_{q,t},\qquad
b_{q,t}=\frac{\Delta_t}{2}\sqrt{Nd},
\label{app3:eq:qat-pointwise-bound}
\end{equation}
where $\Delta_t$ is the current quantization scale and $N,d$ are the aggregation
output dimensions.

During joint QAT, the routing map $\phi_t$, low-rank factors $P_{A,t},P_{B,t}$,
backbone $\Theta_t$, and trainable quantizer parameters are updated together
under the task loss rather than by fixed-$R$ gradient descent as in
\eqref{app3:eq:free-prompt-convergence}. Even under these coupled updates,
\eqref{app3:eq:qat-pointwise-bound} holds pointwise at every iteration $t$,
bounding the post-requantization aggregation error via the current prompt
residual $\varepsilon_t(P_t)$ and scale $\Delta_t$ without freezing the
backbone or adding an auxiliary reconstruction loss.
\end{purplerevisionblock}

\hrulefill
\begin{purplerevisionblock}
\subsection{Layer-wise Analysis of Aggregation Residual Compensation}
\label{app:residual-analysis}

Section~IV of the main paper defines the aggregation residual
$R=S_{\mathrm{fp}}-\widehat S$ and describes how a prompt output $P$ changes
this residual before and after requantization. We measure these quantities
layer by layer in two configurations, each with four message-passing layers and
prompt rank $r=8$: GCN on MNIST under standard QAT and GIN on CIFAR-10 under
A$^2$Q (Fig.~\ref{fig:residual_compensation}). In both rows, panel~(a) reports
the continuous residual ratio $\|R-P\|_F/\|R\|_F$ and the requantized error
ratio $e_q/e_{\mathrm{base}}$, where $e_{\mathrm{base}}$ is the error of the
requantized aggregation without a prompt. Panel~(b) shows the cumulative energy
$E_k=\sum_{j\le k}\sigma_j^2/\sum_j\sigma_j^2$ of the singular values
$\sigma_j$ of $R$, and panel~(c) tracks the continuous ratio during training.

\begin{figure}[!htbp]
\color{theorypurple}
\centering
\includegraphics[width=\linewidth]{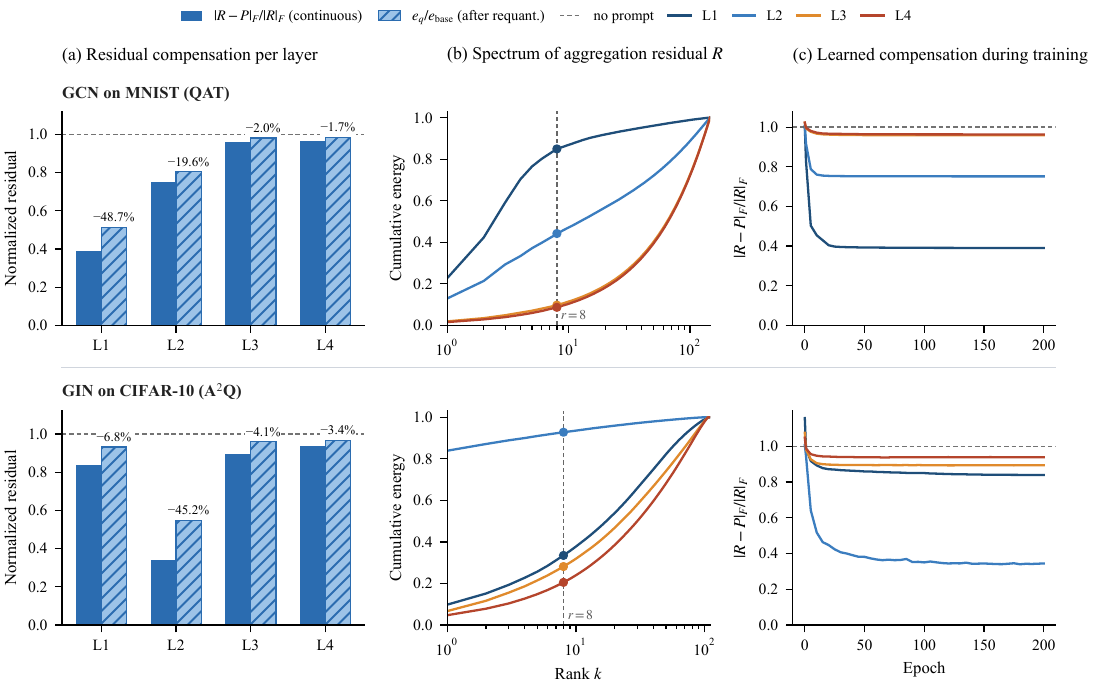}
\caption{Layer-wise compensation of the aggregation residual
$R=S_{\mathrm{fp}}-\widehat S$ by LoRAP ($r=8$) for GCN on MNIST under
standard QAT (top row) and GIN on CIFAR-10 under A$^2$Q (bottom row).
(a)~Continuous residual ratio $\|R-P\|_F/\|R\|_F$ and requantized error ratio
$e_q/e_{\mathrm{base}}$. The dashed line marks the no-prompt value, and the
percentages give the relative reduction $1-e_q/e_{\mathrm{base}}$.
(b)~Cumulative energy $E_k$ of the singular values of $R$; markers indicate $E_8$.
(c)~Continuous residual ratio during training.}
\label{fig:residual_compensation}
\end{figure}

\paragraph{GCN on MNIST under QAT.}
In the top row of Fig.~\ref{fig:residual_compensation}, the prompt reduces the residual in
all four layers, and the reduction persists after requantization. The
continuous ratio is 0.39 in the first layer and 0.75 in the second, and the
requantized error decreases by 48.7\% and 19.6\%, respectively. In the third and
fourth layers, the continuous ratio remains at 0.96 and the requantized error
decreases by about 2\%. Panel~(b) accounts for this difference: $E_8$ is
84.9\% in the first layer and 44.2\% in the second, but below 10\% in the third
and fourth layers, so most of the residual energy in the deeper layers lies
outside any rank-8 subspace. In panel~(c), all ratios start slightly above 1 because
the prompt is randomly initialized and not yet aligned with $R$; the first two
layers then decrease within the first few tens of epochs before settling at the
levels shown in panel~(a).

\paragraph{GIN on CIFAR-10 under A$^2$Q.}
In the bottom row of Fig.~\ref{fig:residual_compensation}, the prompt again reduces the
residual in every layer, but the strongest compensation occurs in the second
layer rather than the first. The second-layer continuous ratio is 0.34, and the
requantized error decreases by 45.2\%. The other three layers keep continuous
ratios between 0.84 and 0.94, with requantized reductions of 3.4--6.8\%.
Panel~(b) shows the same spectral pattern as in the GCN case: the leading
singular direction of the second-layer residual already carries most of its
energy, and $E_8$ reaches 92.8\%, whereas the other layers have $E_8$ between
20.5\% and 33.5\%. In panel~(c), the ratios again start above 1, with the first
layer at about 1.16; the second-layer ratio then shows the largest and most
gradual decrease, while the other layers stabilize early at higher values.

\paragraph{Summary.}
In both configurations, requantization adds the error term $\xi_q(P)$ in the
residual identity of Theorem~\ref{app3:thm:requantization}, so each requantized reduction is smaller than the corresponding
continuous one, yet it remains positive in every layer. The two configurations
also differ in which layer benefits most, the first layer for GCN and the second
for GIN, and in both cases this is the layer whose residual has the highest
$E_8$. LoRAP therefore corrects the low-rank component of the aggregation
residual, independently of layer depth. Where the residual concentrates in a few
directions, the reduction is substantial and survives requantization.
\end{purplerevisionblock}

\hrulefill
\begin{lowrankrevisionblock}
\subsection{Transfer of Theoretical Guarantees to Low-Rank LoRAP}
\label{app:lowrank-transfer}

Sections~A.1--A.3 analyze an unconstrained matrix $P\in\R^{N\times d}$ to isolate the effect of the insertion position. We now establish how optimization geometry, node-level locality, and requantization error preservation transfer to LoRAP's shared low-rank parameterization introduced in Section~V of the main paper.

\paragraph{Parameterized setup.}
For $k$ prompt bases factorized at rank $r<\min(k,d)$, let $\Delta_k=\{\alpha\in\R^{1\times k}:\alpha_m>0,\sum_{m=1}^k\alpha_m=1\}$ denote the open probability simplex, and define the post-aggregation and matched pre-aggregation low-rank prompt generators by
\begin{equation}
\begin{aligned}
P(\theta_P)&=M(\phi)P_A P_B, &
M(\phi)&=\mathrm{Softmax}\bigl(\phi(\widehat S)\bigr)\in\Delta_k^N,\\
P_{\mathrm{pre}}(\theta_P)&=M_{\mathrm{pre}}(\phi_{\mathrm{pre}})P_A P_B, &
M_{\mathrm{pre}}(\phi_{\mathrm{pre}})&=\mathrm{Softmax}\bigl(\phi_{\mathrm{pre}}(X_q)\bigr)\in\Delta_k^N,
\end{aligned}
\label{app:eq:lowrank-generators}
\end{equation}
where $\phi(\widehat s_i)=\widehat s_i W_\phi+b_\phi$ and $\phi_{\mathrm{pre}}(x_{q,j})=x_{q,j}W_{\phi,\mathrm{pre}}+b_{\phi,\mathrm{pre}}$ act row-wise with $W_\phi,W_{\phi,\mathrm{pre}}\in\R^{d\times k}$ and $b_\phi,b_{\phi,\mathrm{pre}}\in\R^{1\times k}$, while $P_A\in\R^{k\times r}$ and $P_B\in\R^{r\times d}$ are the shared low-rank factors collected in $\theta_P=(\phi,P_A,P_B)$. Substituting~\eqref{app:eq:lowrank-generators} into~\eqref{app2:eq:objectives} gives the parameterized reconstruction objectives
\begin{equation}
\widehat{\mathcal L}_{\mathrm{post}}(\phi,P_A,P_B)
=\tfrac12\|R-M(\phi)P_A P_B\|_F^2,\qquad
\widehat{\mathcal L}_{\mathrm{pre}}(\phi_{\mathrm{pre}},P_A,P_B)
=\tfrac12\|R-AM_{\mathrm{pre}}(\phi_{\mathrm{pre}})P_A P_B\|_F^2.
\label{app:eq:lowrank-objectives}
\end{equation}
Because $\widehat S$ and $R=S_{\mathrm{fp}}-\widehat S$ are independent of $\theta_P$, all post-aggregation derivatives below hold identically whether $\widehat S=AX_q$ (Section~A.1) or $\widehat S=DQ(S_q)$ includes an aggregation quantizer prior to prompt addition (Section~V-B of the main paper).

\begin{theorem}[Low-rank optimization geometry and landscape]
\label{app:thm:lowrank-geometry}
Fix $A$, $\widehat S$, $X_q$, and the routing matrices $M=M(\phi)$ and $M_{\mathrm{pre}}=M_{\mathrm{pre}}(\phi_{\mathrm{pre}})$ in~\eqref{app:eq:lowrank-objectives}, and let $\Pi_M=MM^\dagger$ be the orthogonal projector onto $\mathrm{col}(M)$. In column-vectorized factor coordinates, the block Hessians satisfy
\begin{equation}
\begin{aligned}
\nabla^2_{P_A}\widehat{\mathcal L}_{\mathrm{post}}
&=(P_B P_B^\top)\otimes(M^\top M), &
\nabla^2_{P_B}\widehat{\mathcal L}_{\mathrm{post}}
&=I_d\otimes(P_A^\top M^\top M P_A),\\
\nabla^2_{P_A}\widehat{\mathcal L}_{\mathrm{pre}}
&=(P_B P_B^\top)\otimes(M_{\mathrm{pre}}^\top A^\top A M_{\mathrm{pre}}), &
\nabla^2_{P_B}\widehat{\mathcal L}_{\mathrm{pre}}
&=I_d\otimes(P_A^\top M_{\mathrm{pre}}^\top A^\top A M_{\mathrm{pre}} P_A).
\end{aligned}
\label{app:eq:lowrank-hessians}
\end{equation}
Moreover, although $(P_A,P_B)\mapsto\widehat{\mathcal L}_{\mathrm{post}}(\phi,P_A,P_B)$ is nonconvex, every critical point $(P_A,P_B)$ is either a global minimum achieving
\begin{equation}
\min_{P_A\in\R^{k\times r},\,P_B\in\R^{r\times d}}
\widehat{\mathcal L}_{\mathrm{post}}(\phi,P_A,P_B)
=\tfrac12\Bigl(\|R\|_F^2-\sum_{j=1}^{r}\sigma_j^2(\Pi_M R)\Bigr),
\label{app:eq:lowrank-global-min}
\end{equation}
with $\sigma_j(\Pi_M R)=0$ for $j>\min(\mathrm{rank}(M),d)$, or a strict saddle point whose joint Hessian $\nabla^2_{(P_A,P_B)}\widehat{\mathcal L}_{\mathrm{post}}$ has a strictly negative eigenvalue.
\end{theorem}

\begin{proof}
For fixed $M$ and $P_B$, differentiating $\widehat{\mathcal L}_{\mathrm{post}}$ along $H_A\in\R^{k\times r}$ gives
\[
D_{P_A}\widehat{\mathcal L}_{\mathrm{post}}[H_A]
=\langle M P_A P_B-R,\,M H_A P_B\rangle_F
=\langle M^\top(M P_A P_B-R)P_B^\top,\,H_A\rangle_F,
\]
and a second derivative yields the linear operator $H_A\mapsto(M^\top M)H_A(P_B P_B^\top)$. Applying $\vecop(UXV)=(V^\top\otimes U)\vecop(X)$ with symmetric $P_B P_B^\top$ gives $\nabla^2_{P_A}\widehat{\mathcal L}_{\mathrm{post}}=(P_B P_B^\top)\otimes(M^\top M)$. Similarly, differentiating with respect to $P_B$ along $H_B\in\R^{r\times d}$ yields $H_B\mapsto(P_A^\top M^\top M P_A)H_B$, whose column vectorization is $I_d\otimes(P_A^\top M^\top M P_A)$. Replacing $M$ with $AM_{\mathrm{pre}}$ gives the two pre-aggregation block Hessians in~\eqref{app:eq:lowrank-hessians}.

To analyze the global landscape of $\widehat{\mathcal L}_{\mathrm{post}}(\phi,P_A,P_B)$ for fixed $M\in\Delta_k^N$, note first that $M\mathbf{1}_k=\mathbf{1}_N\ne0$, so $m:=\mathrm{rank}(M)\ge1$. Let $M=U_M\Sigma_M V_M^\top$ be the compact singular value decomposition of $M$, with $U_M\in\R^{N\times m}$ ($U_M^\top U_M=I_m$), $\Sigma_M\in\R^{m\times m}$ positive diagonal, and $V_M\in\R^{k\times m}$ ($V_M^\top V_M=I_m$). Then $\Pi_M=U_M U_M^\top$. Setting $R_M:=U_M^\top R\in\R^{m\times d}$ and $\widetilde U:=\Sigma_M V_M^\top P_A\in\R^{m\times r}$, we have $M P_A P_B=U_M\widetilde U P_B\in\mathrm{col}(U_M)$ and $(I_N-\Pi_M)R\perp\mathrm{col}(U_M)$, so
\begin{equation}
\widehat{\mathcal L}_{\mathrm{post}}(\phi,P_A,P_B)
=\tfrac12\|(I_N-\Pi_M)R\|_F^2+\tfrac12\|R_M-\widetilde U P_B\|_F^2.
\label{app:eq:projected-factorization}
\end{equation}
Because $P_A\mapsto\Sigma_M V_M^\top P_A$ maps $\R^{k\times r}$ surjectively onto $\R^{m\times r}$ (with right inverse $\widetilde U\mapsto V_M\Sigma_M^{-1}\widetilde U$) and $\sigma_j(R_M)=\sigma_j(\Pi_M R)$ for all $j\ge1$, the Eckart--Young--Mirsky theorem applied to $R_M$ gives
\[
\min_{P_A,P_B}\tfrac12\|R_M-\widetilde U P_B\|_F^2
=\tfrac12\Bigl(\|R_M\|_F^2-\sum_{j=1}^{r}\sigma_j^2(R_M)\Bigr).
\]
Combining this identity with $\|(I_N-\Pi_M)R\|_F^2=\|R\|_F^2-\|R_M\|_F^2$ proves~\eqref{app:eq:lowrank-global-min}.

Now let $(P_A,P_B)$ be any critical point of $\widehat{\mathcal L}_{\mathrm{post}}(\phi,\cdot,\cdot)$ that is not a global minimum, and set $E_M:=R_M-\widetilde U P_B$. Stationarity reads
\[
0=\nabla_{P_A}\widehat{\mathcal L}_{\mathrm{post}}
=-V_M\Sigma_M E_M P_B^\top,\qquad
0=\nabla_{P_B}\widehat{\mathcal L}_{\mathrm{post}}
=-\widetilde U^\top E_M.
\]
Left-multiplying the first equation by $\Sigma_M^{-1}V_M^\top$ gives $E_M P_B^\top=0$, alongside $\widetilde U^\top E_M=0$. Hence $\mathrm{col}(E_M)\perp\mathrm{col}(\widetilde U)\supseteq\mathrm{col}(\widetilde U P_B)$ and $\mathrm{row}(E_M)\perp\mathrm{row}(P_B)\supseteq\mathrm{row}(\widetilde U P_B)$, so the nonzero singular values of $R_M=\widetilde U P_B+E_M$ are the disjoint multiset union of the nonzero singular values of $\widetilde U P_B$ and those of $E_M$. Since $(P_A,P_B)$ does not attain the global minimum in~\eqref{app:eq:lowrank-global-min}, we have $E_M\ne0$ (so $\sigma_1(E_M)>0$) and $\sigma_r(\widetilde U P_B)<\sigma_1(E_M)$, where $\sigma_r(\widetilde U P_B):=0$ if $\mathrm{rank}(\widetilde U P_B)<r$.

Let $u_e\in\R^m$ and $v_e\in\R^d$ be unit top left and right singular vectors of $E_M$, so that $u_e^\top E_M v_e=\sigma_1(E_M)>0$, $\widetilde U^\top u_e=0$, and $P_B v_e=0$. For any $a,b\in\R^r$, define the perturbation $\Delta_{\widetilde U}=u_e a^\top$, $\Delta_{P_A}=V_M\Sigma_M^{-1}\Delta_{\widetilde U}$, and $\Delta_{P_B}=b v_e^\top$. Substituting $(P_A+t\Delta_{P_A},P_B+t\Delta_{P_B})$ into~\eqref{app:eq:projected-factorization} and using $\langle\widetilde U b v_e^\top,u_e a^\top P_B\rangle_F=(u_e^\top\widetilde U b)(a^\top P_B v_e)=0$ yields the second directional derivative
\[
D^2_{(P_A,P_B)}\widehat{\mathcal L}_{\mathrm{post}}[\Delta_{P_A},\Delta_{P_B}]
=\|\widetilde U b\|_2^2+\|P_B^\top a\|_2^2-2\sigma_1(E_M)\,a^\top b.
\]
If $\ker(\widetilde U)\ne\{0\}$, choosing a unit vector $b\in\ker(\widetilde U)$ and $a=\tau b$ with $0<\tau<2\sigma_1(E_M)/(\|P_B^\top b\|_2^2+1)$ makes this quadratic form strictly negative. Symmetrically, if $\ker(P_B^\top)\ne\{0\}$, choosing unit $a\in\ker(P_B^\top)$ and $b=\tau a$ with $0<\tau<2\sigma_1(E_M)/(\|\widetilde U a\|_2^2+1)$ makes it strictly negative. Otherwise, both $\widetilde U^\top\widetilde U$ and $P_B P_B^\top$ are positive definite. Let $w\in\R^r$ be a unit eigenvector associated with the smallest eigenvalue of $(\widetilde U^\top\widetilde U)^{1/2}(P_B P_B^\top)(\widetilde U^\top\widetilde U)^{1/2}$, which equals $\sigma_r^2(\widetilde U P_B)$, and set $a=(\widetilde U^\top\widetilde U)^{1/2}w$ and $b=\sigma_1(E_M)(\widetilde U^\top\widetilde U)^{-1/2}w$. Then $(\Delta_{P_A},\Delta_{P_B})\ne(0,0)$ and
\[
D^2_{(P_A,P_B)}\widehat{\mathcal L}_{\mathrm{post}}[\Delta_{P_A},\Delta_{P_B}]
=\sigma_1^2(E_M)+\sigma_r^2(\widetilde U P_B)-2\sigma_1^2(E_M)
=\sigma_r^2(\widetilde U P_B)-\sigma_1^2(E_M)<0.
\]
By the Rayleigh quotient characterization, $\nabla^2_{(P_A,P_B)}\widehat{\mathcal L}_{\mathrm{post}}$ has a strictly negative eigenvalue.
\end{proof}

In factor coordinates, the Hessian is no longer the identity because curvature depends on the routing Gram matrix $M^\top M$ and factor norms. Nevertheless, Theorem~\ref{app:thm:lowrank-geometry} preserves the structural advantage of Theorem~\ref{app2:thm:geometry}: post-aggregation factor curvature is independent of $A$ and free of spurious local minima for fixed routing, whereas pre-aggregation replaces $M^\top M$ with $M_{\mathrm{pre}}^\top A^\top A M_{\mathrm{pre}}$, inheriting spectral distortion from $A^\top A$ and flat directions whenever $\mathrm{col}(M_{\mathrm{pre}})\cap\ker(A)\ne\{0\}$. When a pre-aggregation prompt is injected prior to input quantization as $A\,DQ(Q(X+P_{\mathrm{pre}}(\theta_P)))$ (as in GPF-plus), QAT evaluates backward derivatives via the straight-through estimator (STE). On entries within the quantizer's clipping interval, the STE Jacobian of $DQ(Q(\cdot))$ is the identity, so the Gauss--Newton curvature of the low-rank factors still replaces $M^\top M$ with $M_{\mathrm{pre}}^\top A^\top A M_{\mathrm{pre}}$, while clipped entries introduce additional zero-gradient directions.

\begin{theorem}[Low-rank locality and residual attribution]
\label{app:thm:lowrank-locality}
For fixed parameters $\theta_P$, let $\alpha_i\in\Delta_k$ denote the $i$th row of $M(\phi)$, $\Sigma_{\alpha,i}=\mathrm{Diag}(\alpha_i)-\alpha_i^\top\alpha_i$, and $J_{\mathrm{prompt},i}=P_B^\top P_A^\top\Sigma_{\alpha,i}W_\phi^\top$. For $S_{\mathrm{post}}(\theta_P)=\widehat S+P(\theta_P)$ and $S_{\mathrm{pre}}(\theta_P)=A(X_q+P_{\mathrm{pre}}(\theta_P))$, every pair of nodes $i,j$ satisfies
\begin{equation}
\frac{\partial(S_{\mathrm{post}}(\theta_P))_i}{\partial\widehat s_j}
=\delta_{ij}(I_d+J_{\mathrm{prompt},i}),\qquad
\frac{\partial(S_{\mathrm{pre}}(\theta_P))_i}{\partial x_{q,j}}
=A_{ij}(I_d+J_{\mathrm{prompt},j}^{\mathrm{pre}}),
\label{app:eq:lowrank-locality}
\end{equation}
where $J_{\mathrm{prompt},j}^{\mathrm{pre}}=P_B^\top P_A^\top\Sigma_{\alpha,j}^{\mathrm{pre}}W_{\phi,\mathrm{pre}}^\top$. Writing $E_i^{\mathrm{post}}=P_i(\theta_P)-R_i$ and $\widetilde E_j^{\mathrm{pre}}=\sum_{i=1}^{N}A_{ij}\bigl((AP_{\mathrm{pre}}(\theta_P))_i-R_i\bigr)$, the shared-parameter gradients satisfy
\begin{equation}
\begin{aligned}
\nabla_{P_A}\widehat{\mathcal L}_{\mathrm{post}}
&=\sum_{i=1}^{N}\alpha_i^\top E_i^{\mathrm{post}}P_B^\top, &
\nabla_{P_B}\widehat{\mathcal L}_{\mathrm{post}}
&=\sum_{i=1}^{N}P_A^\top\alpha_i^\top E_i^{\mathrm{post}}, &
\nabla_{W_\phi}\widehat{\mathcal L}_{\mathrm{post}}
&=\sum_{i=1}^{N}\widehat s_i^\top E_i^{\mathrm{post}}P_B^\top P_A^\top\Sigma_{\alpha,i},\\
\nabla_{P_A}\widehat{\mathcal L}_{\mathrm{pre}}
&=\sum_{j=1}^{N}(\alpha_j^{\mathrm{pre}})^\top\widetilde E_j^{\mathrm{pre}}P_B^\top, &
\nabla_{P_B}\widehat{\mathcal L}_{\mathrm{pre}}
&=\sum_{j=1}^{N}P_A^\top(\alpha_j^{\mathrm{pre}})^\top\widetilde E_j^{\mathrm{pre}}, &
\nabla_{W_{\phi,\mathrm{pre}}}\widehat{\mathcal L}_{\mathrm{pre}}
&=\sum_{j=1}^{N}x_{q,j}^\top\widetilde E_j^{\mathrm{pre}}P_B^\top P_A^\top\Sigma_{\alpha,j}^{\mathrm{pre}}.
\end{aligned}
\label{app:eq:lowrank-param-gradients}
\end{equation}
\end{theorem}

\begin{proof}
For row $i$, $(S_{\mathrm{post}}(\theta_P))_i=\widehat s_i+\alpha_i(\widehat s_i W_\phi+b_\phi)P_A P_B$ depends only on $\widehat s_i$. For $z_i=\widehat s_i W_\phi+b_\phi$, the Softmax derivative is $\partial\alpha_{i,m}/\partial z_{i,p}=(\Sigma_{\alpha,i})_{m,p}$, so for entries $(c,q)\in\{1,\dots,d\}^2$,
\[
\frac{\partial P_{i,c}(\theta_P)}{\partial\widehat s_{j,q}}
=\delta_{ij}\sum_{m=1}^{k}\sum_{p=1}^{k}(W_\phi)_{q,p}(\Sigma_{\alpha,i})_{p,m}(P_A P_B)_{m,c}
=\delta_{ij}(P_B^\top P_A^\top\Sigma_{\alpha,i}W_\phi^\top)_{c,q}
=\delta_{ij}(J_{\mathrm{prompt},i})_{c,q},
\]
which gives the post-aggregation block in~\eqref{app:eq:lowrank-locality}. Similarly, $(S_{\mathrm{pre}}(\theta_P))_i=\sum_{\ell=1}^N A_{i\ell}(x_{q,\ell}+P_{\mathrm{pre},\ell}(\theta_P))$ yields $A_{ij}(I_d+J_{\mathrm{prompt},j}^{\mathrm{pre}})$.

For the parameter gradients, Theorem~\ref{thm:insertion_locality} gives the output-row derivatives $\partial\mathcal L_{\mathrm{post}}/\partial P_i=E_i^{\mathrm{post}}$ and $\partial\mathcal L_{\mathrm{pre}}/\partial P_{\mathrm{pre},j}=\widetilde E_j^{\mathrm{pre}}$. Applying the chain rule to $P_i(\theta_P)=\alpha_i(\widehat s_i W_\phi+b_\phi)P_A P_B$ across $i=1,\dots,N$ gives
\[
\frac{\partial\widehat{\mathcal L}_{\mathrm{post}}}{\partial(P_A)_{m,u}}
=\sum_{i=1}^{N}\alpha_{i,m}\bigl(E_i^{\mathrm{post}}P_B^\top\bigr)_u,\qquad
\frac{\partial\widehat{\mathcal L}_{\mathrm{post}}}{\partial(P_B)_{u,c}}
=\sum_{i=1}^{N}(\alpha_i P_A)_u E_{i,c}^{\mathrm{post}},
\]
and
\[
\frac{\partial\widehat{\mathcal L}_{\mathrm{post}}}{\partial(W_\phi)_{q,p}}
=\sum_{i=1}^{N}\widehat s_{i,q}\sum_{m=1}^{k}\bigl(E_i^{\mathrm{post}}P_B^\top P_A^\top\bigr)_m(\Sigma_{\alpha,i})_{m,p},
\]
which assemble into the post-aggregation identities in~\eqref{app:eq:lowrank-param-gradients}. Replacing $P_i(\theta_P)$, $\widehat s_i$, $\alpha_i$, and $E_i^{\mathrm{post}}$ with $P_{\mathrm{pre},j}(\theta_P)$, $x_{q,j}$, $\alpha_j^{\mathrm{pre}}$, and $\widetilde E_j^{\mathrm{pre}}$ yields the pre-aggregation gradients.
\end{proof}

Shared prompt parameters aggregate gradients across all $N$ nodes, yet Theorem~\ref{app:thm:lowrank-locality} shows that post-aggregation LoRAP pairs each node's feature $\widehat s_i$ and routing weight $\alpha_i$ solely with its own unmixed receiver residual $E_i^{\mathrm{post}}$, while keeping the node-to-node sensitivity block-diagonal in~\eqref{app:eq:lowrank-locality}. Pre-aggregation low-rank prompting instead replaces $E_j^{\mathrm{post}}$ with the graph-mixed residual $\widetilde E_j^{\mathrm{pre}}$, allowing opposite-sign residuals across receiving neighbors to cancel before updating $\theta_P$.

\begin{theorem}[Low-rank approximation and requantization bound]
\label{app:thm:lowrank-requantization}
Let $V_r\in\R^{d\times r}$ collect orthonormal right singular vectors for the top $r$ singular values of $R$, and set $C_r=RV_r\in\R^{N\times r}$ and $B_{\max}(\theta_P)=\max_{1\le m\le k}\|(P_A P_B)_m\|_2$. Then every $\theta_P$ satisfies
\begin{equation}
\varepsilon(P(\theta_P))^2
\ge\max\Biggl\{\sum_{j>r}\sigma_j^2(R),\;
\sum_{i=1}^{N}\bigl(\max\{0,\|R_i\|_2-B_{\max}(\theta_P)\}\bigr)^2\Biggr\},
\label{app:eq:lowrank-lower-bound}
\end{equation}
and setting $P_B=V_r^\top$ gives the exact decomposition
\begin{equation}
\varepsilon(\phi,P_A,V_r^\top)^2
=\sum_{j>r}\sigma_j^2(R)+\|C_r-M(\phi)P_A\|_F^2.
\label{app:eq:lowrank-pythagoras}
\end{equation}
\begin{secondlowrankrevisionblock}%
For unconstrained simplex weights, if $k\le r$, then either $\mathrm{rank}(P_A)<r$ or $\mathrm{conv}(P_A)$ is confined to an open half-space in $\R^r$. Conversely, whenever $k\ge r+1$, there exist $P_A\in\R^{k\times r}$ of rank $r$ and an unconstrained $M^*\in\Delta_k^N$ (which need not be realizable by the affine router $M(\phi)$) such that $M^*P_A=C_r$, giving $\varepsilon(\phi,P_A,V_r^\top)^2=\sum_{j>r}\sigma_j^2(R)+\|(M(\phi)-M^*)P_A\|_F^2$.%
\end{secondlowrankrevisionblock}
Finally, if post-insertion requantization uses nearest rounding without clipping and entrywise positive scales $\Delta_{i,c}>0$, then with $b_q:=\tfrac12\sqrt{\sum_{i=1}^N\sum_{c=1}^d\Delta_{i,c}^2}$,
\begin{equation}
|e_q(P(\theta_P))-\varepsilon(P(\theta_P))|\le b_q,\qquad
e_q(P(\theta_P))\le\varepsilon(P(\theta_P))+b_q,
\label{app:eq:lowrank-requant}
\end{equation}
so $e_q(P(\theta_P))<e_{\mathrm{base}}$ holds whenever $\varepsilon(P(\theta_P))+b_q<e_{\mathrm{base}}$.
\end{theorem}

\begin{proof}
Because $\mathrm{rank}(P(\theta_P))\le r$, the Eckart--Young--Mirsky theorem gives $\varepsilon(P(\theta_P))^2\ge\sum_{j>r}\sigma_j^2(R)$. In addition, each row $\alpha_i\in\Delta_k$ satisfies $\|P_i(\theta_P)\|_2=\|\sum_{m=1}^k\alpha_{i,m}(P_A P_B)_m\|_2\le B_{\max}(\theta_P)$, so $\|R_i-P_i(\theta_P)\|_2\ge\max\{0,\|R_i\|_2-B_{\max}(\theta_P)\}$, and summing the squared bounds over $i=1,\dots,N$ proves~\eqref{app:eq:lowrank-lower-bound}.

Next, since $V_r^\top V_r=I_r$ and $C_r=RV_r$, we have $(R-C_r V_r^\top)V_r=0$ and $\|R-C_r V_r^\top\|_F^2=\sum_{j>r}\sigma_j^2(R)$. Writing $R-M(\phi)P_A V_r^\top=(R-C_r V_r^\top)+(C_r-M(\phi)P_A)V_r^\top$, the Frobenius inner product of the two terms vanishes:
\[
\langle R-C_r V_r^\top,(C_r-M(\phi)P_A)V_r^\top\rangle_F
=\langle(R-C_r V_r^\top)V_r,C_r-M(\phi)P_A\rangle_F=0.
\]
Using $\|(C_r-M(\phi)P_A)V_r^\top\|_F^2=\|C_r-M(\phi)P_A\|_F^2$ yields~\eqref{app:eq:lowrank-pythagoras}.

For the basis-count threshold, let $q_1,\dots,q_k\in\R^{1\times r}$ denote the rows of $P_A$ and $K=\mathrm{conv}(\{q_1,\dots,q_k\})$. Suppose $k\le r$. If $0\in K$, then $\sum_{m=1}^k\lambda_m q_m=0$ for some $\lambda_m\ge0$ with $\sum_{m=1}^k\lambda_m=1$, so $q_1,\dots,q_k$ are linearly dependent and $\mathrm{rank}(P_A)\le k-1<r$. If $0\notin K$, since $\{0\}$ and $K$ are disjoint compact convex sets, the strict separating hyperplane theorem gives a unit vector $u\in\R^{1\times r}$ such that $q u^\top<0$ for all $q\in K$, confining $K$ to an open half-space. Conversely, suppose $k\ge r+1$ and let $\bar\rho=\max\{1,\max_{1\le i\le N}\|(C_r)_i\|_\infty\}>0$. Define the rows of $P_A\in\R^{k\times r}$ by $q_m=2r(r+1)\bar\rho\,e_m$ for $m=1,\dots,r$ (where $e_m$ is the $m$th standard basis row) and $q_m=-2(r+1)\bar\rho\,\mathbf{1}_r^\top$ for $m=r+1,\dots,k$. The leading $r\times r$ block of $P_A$ is $2r(r+1)\bar\rho\,I_r$, so $\mathrm{rank}(P_A)=r$. For each node $i$ with $c_i=(C_r)_i$, set
\[
\beta_i=\frac12\left(1-\frac{\sum_{m=1}^r c_{i,m}}{2r(r+1)\bar\rho}\right),\qquad
\alpha_{i,m}^*=
\begin{cases}
\dfrac{c_{i,m}+2(r+1)\bar\rho\,\beta_i}{2r(r+1)\bar\rho}, & 1\le m\le r,\\[6pt]
\dfrac{\beta_i}{k-r}, & r+1\le m\le k.
\end{cases}
\]
Because $|c_{i,m}|\le\bar\rho$, we have $\beta_i\ge(2r+1)/[4(r+1)]>0$ and $c_{i,m}+2(r+1)\bar\rho\,\beta_i\ge\bigl(r-\tfrac12\bigr)\bar\rho>0$, so $\alpha_{i,m}^*>0$ for all $m$, $\sum_{m=1}^k\alpha_{i,m}^*=1$, and $\sum_{m=1}^k\alpha_{i,m}^*q_m=c_i$.
\begin{secondlowrankrevisionblock}%
Stacking $\alpha_i^*$ as rows gives an unconstrained $M^*\in\Delta_k^N$ with $M^*P_A=C_r$, and substituting $C_r=M^*P_A$ into~\eqref{app:eq:lowrank-pythagoras} yields $\varepsilon(\phi,P_A,V_r^\top)^2=\sum_{j>r}\sigma_j^2(R)+\|(M(\phi)-M^*)P_A\|_F^2$. Because $M(\phi)=\mathrm{Softmax}(\widehat S W_\phi+b_\phi)$ has logits affine in $\widehat S$, $M^*$ need not be realizable by $\phi$ (for instance, when distinct full-precision neighborhoods quantize to identical rows $\widehat s_i=\widehat s_j$ while $(C_r)_i\ne(C_r)_j$), so the lower bound $\sum_{j>r}\sigma_j^2(R)$ characterizes the capacity of unconstrained convex combinations rather than a guaranteed error of LoRAP.%
\end{secondlowrankrevisionblock}

Finally, under nearest rounding without clipping at entrywise scales $\Delta_{i,c}>0$, each entry of $\xi_q(P(\theta_P))=DQ(Q(Z_{P(\theta_P)}))-Z_{P(\theta_P)}$ satisfies $|(\xi_q(P(\theta_P)))_{i,c}|\le\Delta_{i,c}/2$, so $\|\xi_q(P(\theta_P))\|_F\le b_q$. Applying the triangle and reverse triangle inequalities to $S_{\mathrm{fp}}-\widetilde S(P(\theta_P))=R-P(\theta_P)-\xi_q(P(\theta_P))$ proves~\eqref{app:eq:lowrank-requant}.
\end{proof}

\paragraph{Interpretation and consistency with the layer-wise spectral measurements.}
\begin{secondlowrankrevisionblock}%
Theorem~\ref{app:thm:lowrank-requantization} transfers Theorem~\ref{app3:thm:requantization} and Corollaries~\ref{app3:cor:propagation}--\ref{app3:cor:margin} to low-rank LoRAP conditionally on $\varepsilon(P(\theta_P))+b_q<e_{\mathrm{base}}$, while explaining the design constraint $r<k$: at least $r+1$ bases are necessary for convex combinations of the prompt bases to span both signs across $r$ principal directions, whereas the remaining gap $\|(M(\phi)-M^*)P_A\|_F$ reflects the predictability of $C_r$ from $\widehat S$ under the affine router $\phi$.%
\end{secondlowrankrevisionblock}
When $R\ne0$, the spectral lower bound in~\eqref{app:eq:lowrank-lower-bound} implies $\varepsilon(P(\theta_P))/\|R\|_F\ge\sqrt{1-E_r}$, where $E_r$ is the cumulative singular-value energy defined in Section~\ref{app:residual-analysis}. At $r=8$, the empirical values in Fig.~\ref{fig:residual_compensation} closely track this bound. For GCN on MNIST under QAT, $E_8=84.9\%$ in layer~1, $44.2\%$ in layer~2, and $<10\%$ in layers~3--4 give theoretical floors $\sqrt{1-E_8}\approx0.39$, $0.75$, and $>0.95$, matching the observed continuous ratios $0.39$, $0.75$, and $0.96$. For GIN on CIFAR-10 under A$^2$Q, layer~2 reaches $E_8=92.8\%$ ($\sqrt{1-E_8}\approx0.27$, observed $0.34$), whereas the remaining layers have $E_8\in[20.5\%,33.5\%]$ (floors $0.82$--$0.89$, observed $0.84$--$0.94$). Setting $\Delta_{i,c}\equiv\Delta$ recovers $b_q=\tfrac{\Delta}{2}\sqrt{Nd}$ in Theorem~\ref{app3:thm:requantization} and the pointwise joint-QAT bound~\eqref{app3:eq:qat-pointwise-bound}, while heterogeneous $\Delta_{i,c}$ extends both bounds directly to per-channel quantization and nodewise mixed-precision schemes such as A$^2$Q.
\end{lowrankrevisionblock}

\begin{purplerevisionblock}
\section{Additional Experimental Details and Results}
\label{app:additional-experiments}

\subsection{Adaptation of GraphLoRA to QAT}
\label{app:graphlora}
To compare LoRAP against weight-space low-rank adaptation under quantization, we adapt GraphLoRA~\cite{graphlora2024} (reported as \emph{GraphLoRA} in Tables~I and~V of the main paper) into a parallel low-rank graph adapter within each message-passing layer $\ell\in\{1,\dots,L\}$ of the quantized backbone:
\begin{equation}
Y_\ell = F_\ell^{\mathrm{QAT}}\bigl(\widetilde H_{\ell-1}, \mathcal{E};\, W_\ell, \theta_{Q,\ell}\bigr),
\qquad
\widetilde Y_\ell = Y_\ell + G_\ell\bigl(U_\ell, \mathcal{E};\, \psi_\ell\bigr),
\label{app:eq:graphlora-layer}
\end{equation}
where $F_\ell^{\mathrm{QAT}}$ denotes the native quantized message-passing module with backbone weights $W_\ell$ and quantizer configuration $\theta_{Q,\ell}$, retaining the host framework's fake-quantization operators and backward rules. Here $U_\ell$ is an intermediate tensor reused from the native propagation path without introducing extra quantizer calls: for GIN and GCN, $U_\ell = \widehat S_\ell$ is the dequantized neighborhood aggregate, whereas for GAT, $U_\ell$ is the layer's native propagation input paired with the edge set $\mathcal{E}$ to form a separate low-rank attention branch. The adapter $G_\ell(\cdot;\,\psi_\ell)$ applies an FP32 low-rank projection $W_{A,\ell}W_{B,\ell}$ ($W_{A,\ell}\in\R^{d_{\ell,\mathrm{in}}\times r_{\mathrm{gl}}}$, $W_{B,\ell}\in\R^{r_{\mathrm{gl}}\times d_{\ell,\mathrm{out}}}$) in row-wise node coordinates, with the output factor $W_{B,\ell}$ initialized to zero so that the adapted layer initially matches the unadapted quantized backbone. The sum $\widetilde Y_\ell$ is then processed by the layer's remaining native operations (such as normalization, activation, or skip connections) to produce $\widetilde H_\ell$. Adapter parameters remain in FP32 and are not merged into the quantized weights $W_\ell$.

Unlike the cross-graph transfer setting in~\cite{graphlora2024}, our QAT benchmark operates in a single-domain training regime without source-graph pretraining, SMMD distribution alignment, or auxiliary transfer losses (omitting the contrastive and structural reconstruction terms of~\cite{graphlora2024}). We jointly optimize the backbone weights $W$, quantizer parameters $\theta_Q$ (when trainable under the host recipe), and adapter parameters $\psi=\{\psi_\ell\}_{\ell=1}^L$ under the host QAT framework's native training objective (the task loss $L_D$, together with the native bit-budget storage penalty of $A^2Q$ when applicable), using the backbone learning rate and zero weight decay for $\psi$.

\subsection{Equal-Capacity MLP and FP32 Prompting Comparison}
\label{app:equal-capacity-mlp}
Table~\ref{tab:equal_capacity_mlp} compares GPF-LoRAP against no prompting and an equal-capacity MLP baseline under $A^2Q$, alongside full-precision (FP32) baselines with and without GPF-LoRAP. Specifically, the equal-capacity MLP replaces LoRAP at each layer with a two layer ReLU residual MLP whose per-layer parameter count matches that of our input-dependent low-rank prompt generator, and is jointly optimized with the backbone and quantizer parameters under the host $A^2Q$ objective. Replacing GPF-LoRAP with this equal-capacity MLP frequently degrades $A^2Q$ accuracy, while applying GPF-LoRAP directly to FP32 models yields no consistent gain and often lowers accuracy, indicating that LoRAP specifically compensates for low-bit quantization distortions rather than its extra capacity.

\begin{table}[!htbp]
\color{theorypurple}
\centering
\caption{Test accuracy (\%) comparison under $A^2Q$ (with no prompting, an equal-capacity MLP, and GPF-LoRAP) and FP32 (with and without GPF-LoRAP).}
\label{tab:equal_capacity_mlp}
\begin{tabular}{ll ccc cc}
\toprule
\multirow{2}{*}{\textbf{Dataset}} & \multirow{2}{*}{\textbf{Model}}
& \multicolumn{3}{c}{\textbf{$A^2Q$}}
& \multicolumn{2}{c}{\textbf{FP32}} \\
\cmidrule(lr){3-5} \cmidrule(lr){6-7}
& & \textbf{None} & \textbf{Equal-Capacity MLP} & \textbf{GPF-LoRAP} & \textbf{None} & \textbf{GPF-LoRAP} \\
\midrule
\multirow{3}{*}{Cora}
& GIN & $77.4 \pm 0.9$ & $77.4 \pm 0.5$ & $78.5 \pm 1.0$ & $77.6 \pm 1.1$ & $78.0 \pm 0.7$ \\
& GCN & $76.1 \pm 0.3$ & $76.8 \pm 0.3$ & $79.1 \pm 0.1$ & $81.4 \pm 0.7$ & $77.0 \pm 0.8$ \\
& GAT & $82.6 \pm 0.8$ & $80.9 \pm 0.4$ & $83.6 \pm 0.6$ & $83.1 \pm 0.4$ & $81.0 \pm 1.2$ \\
\midrule
\multirow{3}{*}{CiteSeer}
& GIN & $66.6 \pm 0.1$ & $64.0 \pm 1.6$ & $70.7 \pm 0.6$ & $66.1 \pm 0.9$ & $68.0 \pm 0.7$ \\
& GCN & $69.7 \pm 1.6$ & $67.7 \pm 0.9$ & $71.8 \pm 0.7$ & $71.1 \pm 0.7$ & $66.9 \pm 1.8$ \\
& GAT & $71.2 \pm 0.9$ & $68.9 \pm 1.2$ & $72.5 \pm 0.8$ & $72.5 \pm 0.7$ & $66.8 \pm 3.4$ \\
\midrule
\multirow{3}{*}{PubMed}
& GIN & $72.9 \pm 0.4$ & $73.2 \pm 2.6$ & $75.7 \pm 1.4$ & $75.9 \pm 1.2$ & $75.7 \pm 1.7$ \\
& GCN & $71.4 \pm 0.2$ & $73.3 \pm 1.8$ & $77.3 \pm 1.1$ & $79.5 \pm 0.2$ & $79.4 \pm 0.4$ \\
& GAT & $78.3 \pm 0.4$ & $74.6 \pm 1.6$ & $80.2 \pm 1.2$ & $78.7 \pm 0.2$ & $78.4 \pm 0.2$ \\
\bottomrule
\end{tabular}
\end{table}

\subsection{End-to-End Performance}
\label{app:end-to-end-performance}

\begin{table}[!htbp]
\color{theorypurple}
\centering
\caption{End-to-end full-graph inference latency, peak GPU memory, and model storage (including prompts) on ogbn-arxiv with a 3-layer GCN ($128 \to 256 \to 256 \to 40$, hidden dimension $d=256$) on an NVIDIA RTX 4090 GPU. GPF-LoRAP uses $k=20$ prompts and rank $r=8$.}
\label{tab:e2e_ogbn_arxiv}
\begin{tabular}{llccc}
\toprule
\textbf{Deployment Path} & \textbf{Prompting} & \textbf{GPU Latency (ms)} & \textbf{Peak GPU Memory (MiB)} & \textbf{Model Storage (KiB)} \\
\midrule
FP32        & None                & $5.530 \pm 0.003$ & 610.03 & 434.16 \\
Native W4A4 & None                & $3.175 \pm 0.004$ &  65.15 &  65.31 \\
Native W4A4 & GPF-LoRAP (unfused) & $4.689 \pm 0.011$ &  94.43 & 114.09 \\
Native W4A4 & GPF-LoRAP (fused)   & $3.449 \pm 0.005$ &  73.68 & 114.09 \\
\bottomrule
\end{tabular}
\end{table}

Table~\ref{tab:e2e_ogbn_arxiv} reports end-to-end full-graph inference latency, peak GPU memory, and model storage (including prompts) on ogbn-arxiv for a 3-layer GCN ($128 \to 256 \to 256 \to 40$, hidden dimension $d=256$) on an NVIDIA RTX 4090 GPU, with GPF-LoRAP configured at $k=20$ and $r=8$. Relative to FP32 (5.530~ms), native W4A4 lowers latency to 3.175~ms ($1.74\times$ speedup). Unfused GPF-LoRAP raises latency to 4.689~ms (+1.514~ms, or +47.7\% over native W4A4), whereas kernel fusion lowers it to 3.449~ms (+0.274~ms, or +8.6\% over native W4A4), retaining a $1.60\times$ speedup over FP32. Peak GPU memory follows the same trend: native W4A4 reduces peak memory from 610.03~MiB to 65.15~MiB (an 89.3\% reduction), unfused GPF-LoRAP raises it to 94.43~MiB (+29.28~MiB, or +44.9\%), and fused GPF-LoRAP brings it down to 73.68~MiB (+8.53~MiB, or +13.1\% over native W4A4, and an 87.9\% reduction relative to FP32). Including prompts, model storage increases from 65.31~KiB under native W4A4 to 114.09~KiB (+48.78~KiB), remaining 73.7\% smaller than the 434.16~KiB FP32 model; the added 48.78~KiB is below 0.1\% of the 73.68~MiB peak runtime memory.
\end{purplerevisionblock}

\end{document}